\documentclass[10pt,twocolumn,letterpaper]{article}

\usepackage{cvpr}
\usepackage{times}
\usepackage{epsfig}
\usepackage{graphicx}
\usepackage{amsmath}
\usepackage{amssymb}
\usepackage{times}
\usepackage{epsfig}
\usepackage{graphicx}
\usepackage{amsmath}
\usepackage{amssymb}
\usepackage{subfigure}
\usepackage{multirow}
\usepackage{booktabs}
\usepackage{array}
\usepackage{color}
\usepackage{caption}
\usepackage{stfloats}
\usepackage{cite}
\usepackage{url}
\usepackage{overpic}
\usepackage{xspace}
\usepackage{bm}
\usepackage{rotating}
\usepackage{xfrac}
\usepackage{mathptmx}
\usepackage[symbol]{footmisc}

\usepackage[pagebackref=true,breaklinks=true,letterpaper=true,colorlinks,bookmarks=false]{hyperref}

 \cvprfinalcopy 

\def\cvprPaperID{4832} 

\ifcvprfinal\fi

\newcommand{\PreserveBackslash}[1]{\let\temp=\\#1\let\\=\temp}
\newcolumntype{C}[1]{>{\PreserveBackslash\centering}p{#1}}
\newcolumntype{R}[1]{>{\PreserveBackslash\raggedleft}p{#1}}
\newcolumntype{L}[1]{>{\PreserveBackslash\raggedright}p{#1}}
\newcommand{\tabincell}[2]{\begin{tabular}{@{}#1@{}}#2\end{tabular}}

\begin{document}

\title{Domain-invariant Stereo Matching Networks}

\author{Feihu Zhang$^1$\quad Xiaojuan Qi$^1$ \quad Ruigang Yang$^2$\quad Victor Prisacariu$^1$ \quad Benjamin Wah$^3$\quad Philip Torr$^1$\\
$^1$University of Oxford\qquad $^2$Baidu Research\qquad $^3$CUHK
}

\maketitle

\begin{abstract}
State-of-the-art stereo matching networks have difficulties in generalizing to new unseen environments due to significant domain differences, such as color, illumination, contrast, and texture. In this paper, we aim at designing a domain-invariant stereo matching network (DSMNet) that generalizes well to unseen scenes.
To achieve this goal, we propose i) a novel ``domain normalization'' approach that regularizes the distribution of learned representations to allow them to be invariant to domain differences, and ii) a trainable non-local graph-based filter for extracting robust structural and geometric representations that can further enhance domain-invariant generalizations. When trained on synthetic data and generalized to real test sets, our model performs significantly better than all state-of-the-art models. It even outperforms some deep learning models (\eg MC-CNN \cite{zbontar2015computing}) fine-tuned with test-domain data. {\color{magenta}{\small The code and dataset will be avialable at \url{https://github.com/feihuzhang/DSMNet}.}}

\end{abstract}

\vspace{-1.5mm}
\section{Introduction}

Stereo reconstruction is a fundamental problem in computer vision, robotics and autonomous driving. It aims to estimate 3D geometry by computing disparities between matching pixels in a stereo image pair.
Recently, many end-to-end deep neural network models (\eg \cite{Zhang2019GANet,chang2018pyramid,kendall2017end}) have been developed for stereo matching that achieve impressive accuracy on several datasets or benchmarks.

\begin{figure*}[t]
 \setlength{\abovecaptionskip}{5pt}
\setlength{\belowcaptionskip}{-6pt}
 \vspace{-2mm}
 \centering
 \begin{overpic}[width=0.975\linewidth]{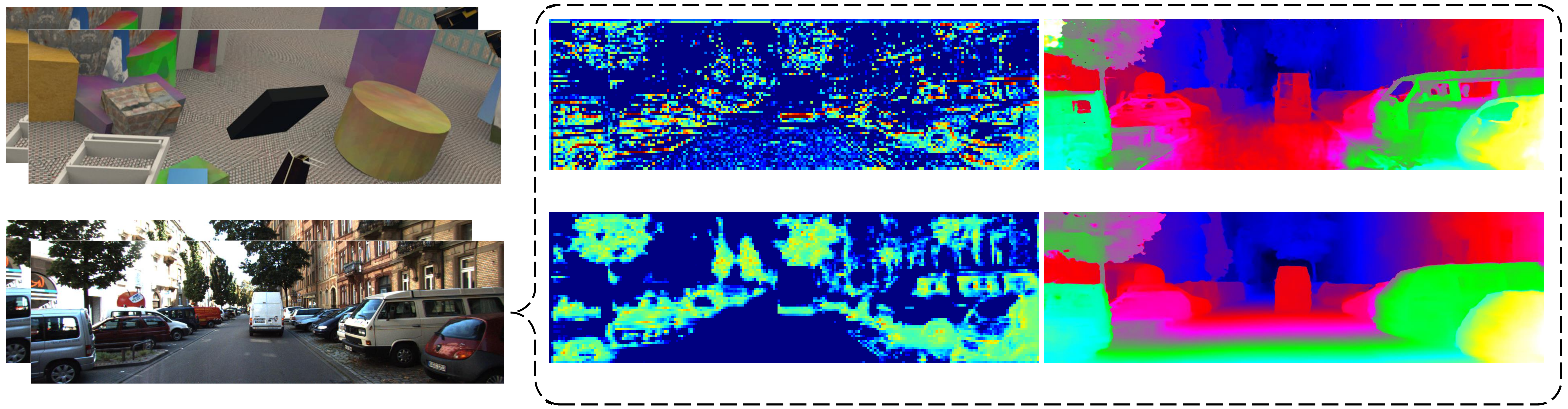}

	    \put(9.5,-0.05){\color{black}{{\small (b) Test Scenes}}}
	   \put(8.5, 12.75){\color{black}{{\small (a) Training Scenes}}}
	    \put(39.5,13.5){\color{black}{{\small (c) Feature Map of GANet \cite{Zhang2019GANet}}}}
	   \put(39, 1){\color{black}{{\small (d) Feature Map of our DSMNet}}}
	    \put(73.5,13.5){\color{black}{{\small (e) Results of GANet \cite{Zhang2019GANet}}}}
	   \put(73.5, 1){\color{black}{{\small (f) Results of Our DSMNet}}}
 \end{overpic}
 \caption{\small Visualization of the feature maps and disparity results. The state-of-the-art GANet \cite{Zhang2019GANet} is used for comparisons. Models are trained on synthetic data (Sceneflow \cite{mayer2016large}) and tested on novel real scenes (KITTI \cite{kitti2015}). 
 The feature maps from GANet has many artifacts (\eg noises).
  Our DSMNet mainly captures the structure and shape information as robust features, and there is no distortions or artifacts in the feature map. It can produce accurate disparity estimations in the novel test scenes. The same observations are shown by more models (\eg PSMNet \cite{chang2018pyramid}, HD$^3$ \cite{yin2019hierarchical}) and datasets in the supplementary material.}
 \label{fig:illustrate}
\end{figure*}
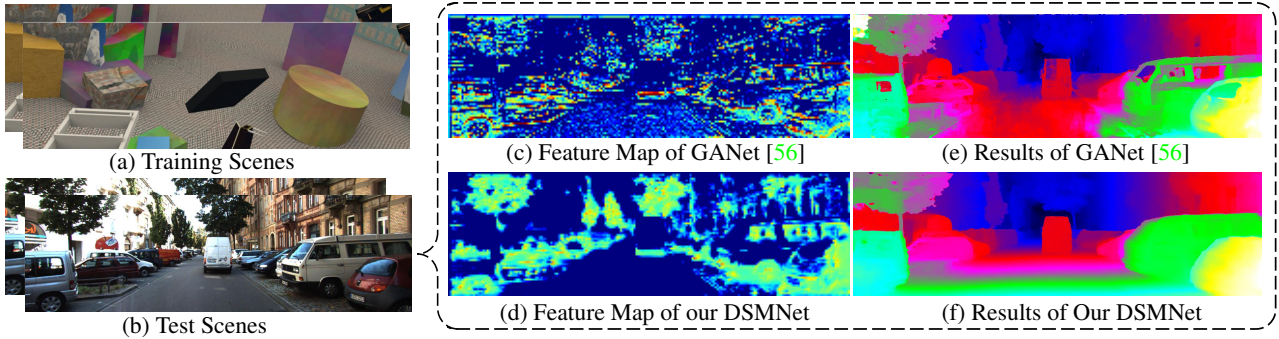

However, state-of-the-art stereo matching networks (supervised \cite{Zhang2019GANet,chang2018pyramid,kendall2017end} and unsupervised \cite{zhou2017unsupervised,tonioni2019real}) cannot generalize well to unseen data without fine-tuning or adaptation. Their difficulties lie in the large domain differences (such as color, illumination, contrast and texture) between stereo images in various datasets. As illustrated in Fig. \ref{fig:illustrate}, the pre-trained models on one specific dataset produce poor results on other real and unseen scenes.

Domain adaptation and transfer learning methods (\eg \cite{tonioni2019real,guo2018learning,bousmalis2017unsupervised}) attempt to transfer or adapt 
from one source domain to another new domain. 
Typically,  a large number of stereo images from the new domain are required for the adaptation. 
However, these cannot be easily obtained in many real scenarios. 
  And, in this case, we still need a good method for disparity estimation even without  data from the new domain for adaptation. 

Thus, it is desirable to design a model that can generalize well to unseen data without re-training or adaptation.
The difficulties for developing such a domain invariant stereo matching network (DSMNet) come from the significant domain differences 
between stereo images in various scenes/datasets (\eg Fig. {\color{red}\ref{fig:illustrate}(a)} and {\ref{fig:illustrate}\color{red}(b)}). Such differences make the learned features unstable, distorted and noisy, leading to many wrong matching results.

Fig. {\ref{fig:illustrate}} visualizes the features learned by some state-of-the-art stereo matching models \cite{Zhang2019GANet,chang2018pyramid,yin2019hierarchical}. Due to the limited effective receptive field of convolutional neural networks \cite{luo2016understanding}, they capture the domain-sensitive local patterns (\eg local contrast, edge and texture) when constructing matching features,
which, however, break down and produce a lot of artifacts (\eg noises) in the feature
maps when applied to the novel test data (Fig. {\color{red}\ref{fig:illustrate}(c)}).  The artifacts and distortions in the features inhibit robust matching, leading to wrong matching results (Fig. {\color{red}\ref{fig:illustrate}(e)}).

In this paper, we propose two novel neural network layers for constructing the robust deep stereo matching network for cross-domain generalization without further fine-tuning or adaptation.
Firstly, to reduce the domain shifts/differences between different datasets/scenes, we propose a novel domain normalization layer that fully regulates the feature's distribution in both the spatial (height and width) and the channel dimensions. Secondly, to eliminate the  artifacts and distortions in the features, we propose a learnable non-local graph-based filtering layer that can capture more robust structural and geometric representations (\eg shape and structure, as illustrated in Fig. {\color{red}\ref{fig:illustrate}(d)}) for domain-invariant stereo matching.

We formulate our method as an end-to-end deep neural network model and train it only with synthetic data. 
In our experiments, 
 without any fine-tuning or adaptation on the real test datasets, our DSMNet far outperforms: 1) almost all state-of-the-art stereo matching models (\eg GANet\cite{Zhang2019GANet}) trained on the same synthetic dataset, 2) most of the  traditional methods (\eg Cosfter filter, SGM \cite{hirschmuller2008stereo} \etal), 3) most of
the unsupervised/self-supervised models trained on the target test domains. Our model even surpasses some of the fine-tuned (on the target domains) supervised deep neural network models (\eg MC-CNN\cite{zbontar2015computing}, content-CNN\cite{luo2016efficient}, DispNetC \cite{mayer2016large} \etal).

\vspace{-0.5mm}
\section{Related Work}
\vspace{-0.5mm}
\subsection{Deep Neural Networks for Stereo Matching}\label{subsec:deepmodel}
In recent years, deep neural networks have seen great success in the task of stereo matching \cite{kendall2017end,chang2018pyramid,Zhang2019GANet}. These models can be categorized into three types:
1) learning better features for traditional stereo matching algorithms, 2) correlation-based end-to-end deep neural networks, 3) cost-volume based stereo matching networks.

In the first category, deep neural networks have been used to compute patch-wise
similarity scores as the matching costs \cite{zhang2018fundamental,zbontar2015computing}. The costs are then fed into the traditional cost aggregation and disparity computation/refinement methods \cite{hirschmuller2008stereo} to get the final disparity maps. The models are, however, limited by
the traditional matching cost aggregation step and often produce wrong predictions in occluded regions, large textureless/reflective regions and around object edges.

DispNetC \cite{mayer2016large}, a typical method in the second category, computes the correlations by warping between stereo views and attempts to predict the per-pixel disparity by minimizing a regression training loss. Many other sate-of-the-art methods, including  iResNet \cite{liang2018learning}, CRL\cite{Pang_2017},  SegStereo \cite{yang2018segstereo},
EdgeStereo \cite{song2019edgestereo}, HD$^3$ \cite{yin2019hierarchical}, and MADNet \cite{tonioni2019real}, are all based on color or feature correlations between the left and right views for disparity estimation.

The recently developed cost-volume based models explicitly
learn feature extraction, cost volume, and regularization
function all end to end. Examples include GC-Net\cite{kendall2017end}, PSM-Net\cite{chang2018pyramid} , StereoNet \cite{stereonet},
AnyNet \cite{wang2018anytime}, GANet \cite{Zhang2019GANet} and EMCUA \cite{nie2019multi}.  They all utilize a similarity cost as the third dimension to build the 4D cost volume in which the real geometric context is maintained.

There are also others that combine the correlation and cost volume strategies (\eg \cite{guo2019group}).

The common feature of these models is that they all require a large number of training samples with ground truth depth/disparity. More importantly, a model trained on one specific domain cannot generalize well to new scenes without fine-tuning or retraining.

\subsection{Adaptation and Self-supervised Learning}
\paragraph{Self-supervised Learning:}
A recent trend of training stereo matching networks in an unsupervised manner relies on image reconstruction losses that are achieved
by warping left and right views \cite{zhou2017unsupervised,zhong2017self}. However, they cannot solve the occlusions and reflective regions where there is no correspondence between the left and the right views. Also, they cannot generalize well to other new domains.
\vspace{-4mm}
\paragraph{Domain Adaptation:}
Some methods pre-train the models on synthetic data and then explore the cross-domain knowledge to adapt  \cite{guo2018learning,pang2018zoom} for a new domain.
Others focus on the online or offline adaptations  \cite{tonioni2019learning,tonioni2019real,Tonioni_2017_ICCV,poggi2019guided}. For example,  MADNet \cite{tonioni2019real} is proposed to adapt the pre-trained model online and in real time. But, it has poor accuracy even after the adaptation.
Moreover, the domain adaptation approaches require a large number of stereo images from the target domain for adaptations. However, these cannot be easily obtained in many real scenarios. 
  And, in this case, we still need a good method for disparity estimation even without  data from the new domain for adaptation.


\subsection{Cross-Domain Generalization}
Different to domain adaptation, domain generalization is a much harder problem that assumes no access to target information for adaptation or fine-tuning.
There are many approaches that explore the idea of domain-invariant feature learning.
Previous approaches focus on developing data-driven strategies to learn invariant features
from different source domains \cite{motiian2017unified,ghifary2015domain,li2017deeper}.
Some recent methods utilize meta-learning that takes variations in multiple source domains to generalize to novel test distributions \cite{balaji2018metareg,li2018learning}.
Other approaches \cite{li2018deep,li2018domain} employ an invariant adversarial network to learn domain-invariant representation/features for image recognition.
Choy \etal \cite{choy2016universal} develop a universal feature learning framework for visual correspondences using deep metric learning.

In contrast to the above approaches, there are methods that try to improve the batch or instance normalization in order to improve the generalization and robustness for style transfer or image recognition \cite{nam2018batch,li2018adaptive,pan2018two}.

In summary, for stereo matching, 
work is seldom done to improve the generalization ability of the end-to-end deep neural network models, especially when developing the domain-invariant stereo matching networks.

\section{Proposed DSMNet}

To overcome the challenges in cross-domain generalization, we develop in the following sections our domain-invariant stereo matching networks.  These include domain normalization to remove the influence of the domain shifts (\eg color, style, illuminance), as well as non-local graph-based filtering and aggregation to capture the non-local structural and geometric context as robust features for domain-invariant stereo reconstruction.
\begin{figure}[t]
 \setlength{\abovecaptionskip}{4pt}
\setlength{\belowcaptionskip}{-2pt}
 \vspace{-1mm}
 \centering
 \includegraphics[width=1\linewidth,height=0.39\linewidth]{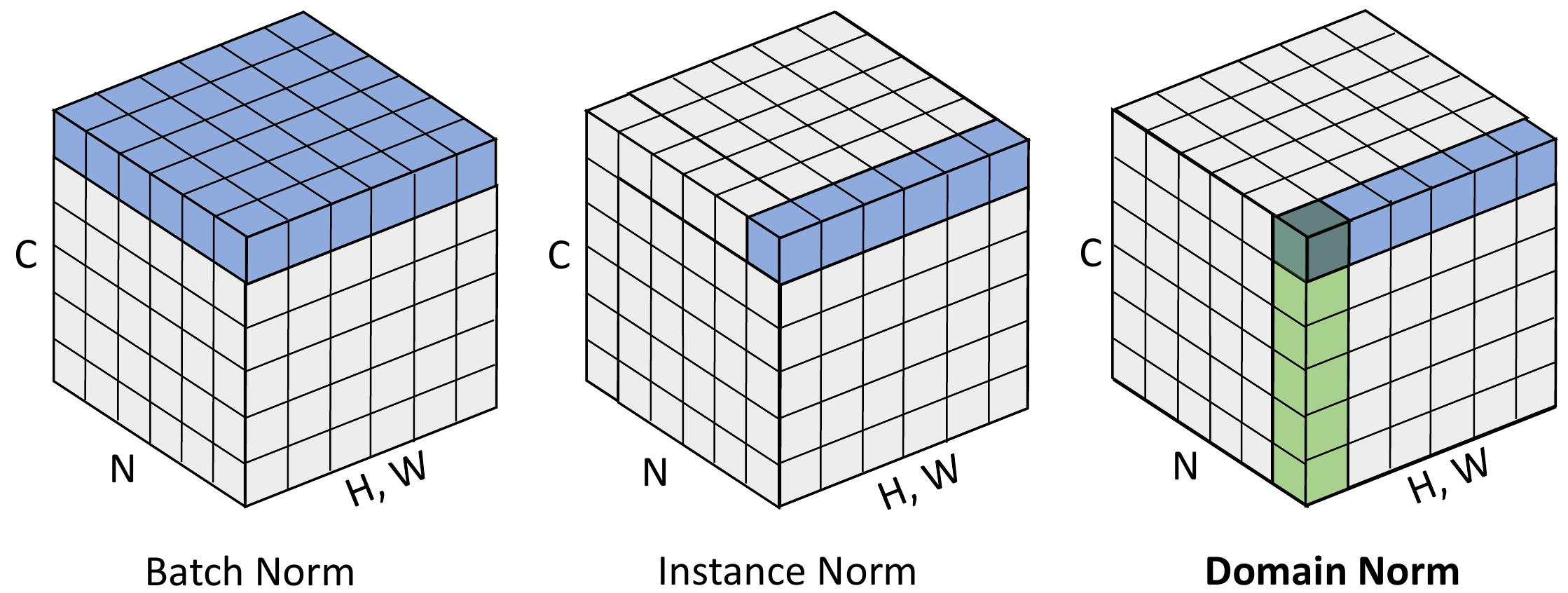}
 \caption{\small Normalization methods. Each subplot shows a feature map tensor, with $N$ as the batch axis, $C$ as the channel axis, and $(H, W)$
as the spatial axes. The blue elements in set $S$ are normalized by the same mean and variance. The proposed domain normalization consists of image-level normalization (blue, Eq.~\eqref{Eq:Norm1}) and pixel-level normalization of each $C$-channel feature vector (green, Eq.~\eqref{Eq:Norm3}). }
 \label{fig:norm}
 \end{figure}
\subsection{Domain Normalization}
\textbf{Batch normalization (BN)} has become the default feature normalization operation for constructing end-to-end deep stereo matching networks~\cite{kendall2017end,chang2018pyramid,Zhang2019GANet,song2019edgestereo,tonioni2019real,mayer2016large}.  Although it can reduce the internal covariate shift effects in training deep networks, it is domain-dependent and has negative influence on the model's cross-domain generalization ability.

BN normalizes the features as follows:
\begin{equation}
\setlength{\abovedisplayskip}{3pt}
\setlength{\belowdisplayskip}{3pt}
 \hat{x}_{i}=\frac{1}{\sigma}(x_i-\mu_i).
\label{Eq:Norm1}
\end{equation}
Here $x$ and $\hat{x}$ are the input and output features, respectively, and $i$ indexes elements in a tensor (\ie feature maps, as illustrated in Fig.~\ref{fig:norm}) of size $N \times C \times H \times W$ ($N$: batch size, $C$: channels, $H$: spatial height, $W$: spatial width).
$\mu_i$ and $\sigma_i$ are the corresponding channel-wise mean and standard deviation (std) and are computed by:
\begin{equation}
\setlength{\abovedisplayskip}{3pt}
\setlength{\belowdisplayskip}{3pt}
\mu_i={\frac {1}{m}}\sum\limits_{k\in S_i}x_{k}, \quad
 \sigma_i=\sqrt{{\frac {1}{m}}\sum\limits_{k\in S_i}(x_{k}-\mu_i)^2+\epsilon},
\label{Eq:Norm2}
\end{equation}
where $S_i$ is the set of elements in the same channel as element $i$ (Fig.~\ref{fig:norm}), and $\epsilon$ is a small constant to avoid dividing by zeros.

Mean $\mu$ and standard deviation $\sigma$ are computed per batch in the training phase, and the accumulated values of the training set are utilized for inference.
However,
different domains may have different $\mu$ and $\sigma$ caused by color shifts, contrast, and illumination (Fig.~{\color{red}\ref{fig:illustrate}(a)} and {\color{red}\ref{fig:illustrate}(b)}).
Thus $\mu$ and $\sigma$ computed for one dataset are not transferable to others.


 \begin{figure}[t]
 \setlength{\abovecaptionskip}{10pt}
\setlength{\belowcaptionskip}{-5pt}
 \vspace{-1mm}
 \centering
 \begin{overpic}[width=0.95\linewidth]{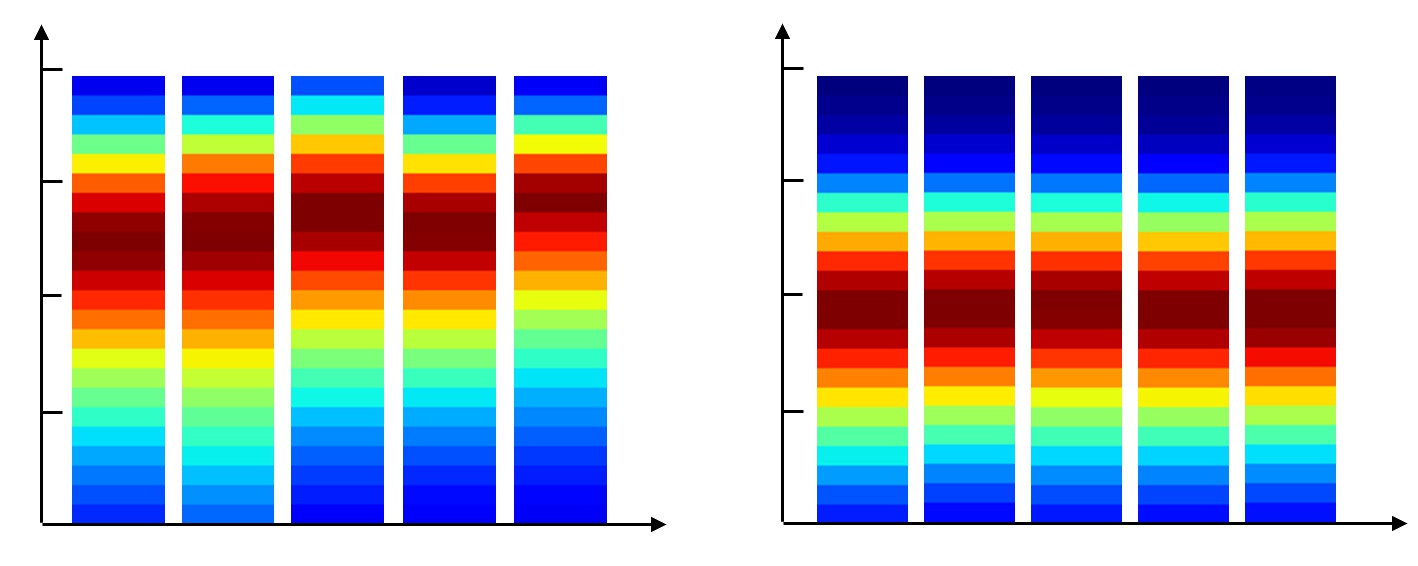}
		\put(-0.25,2){\color{black}{{\footnotesize 1}}}
		\put(-0.25,10){\color{black}{{\footnotesize 2}}}
		\put(-0.25,18){\color{black}{{\footnotesize 3}}}
		\put(-0.25,26){\color{black}{{\footnotesize 4}}}
		\put(-0.25,34){\color{black}{{\footnotesize 5}}}
		\put(52,2){\color{black}{{\footnotesize 1}}}
		\put(52,10){\color{black}{{\footnotesize 2}}}
		\put(52,18){\color{black}{{\footnotesize 3}}}
		\put(52,26){\color{black}{{\footnotesize 4}}}
		\put(52,34){\color{black}{{\footnotesize 5}}}
		\put(10,-1.5){\color{black}{{\small (a) Instance Norm}}}
	   \put(60,-1.5){\color{black}{{\small (b) Domain Norm}}}
 \end{overpic}
 \caption{\small Norm distributions of the features of different datasets (from left to right: synthetic SceneFlow, KITTI, Middlebury, CityScapes  and ETH 3D). 
 We choose the output feature of the feature extraction network for our study. The norm of the $C$-channel feature vector of each pixel is counted for the distribution.  Instance normalization can only reduce the image-level differences, but does not normalize the $C$-channel feature vectors at pixel level.}
 \label{fig:norm2}
 \end{figure}
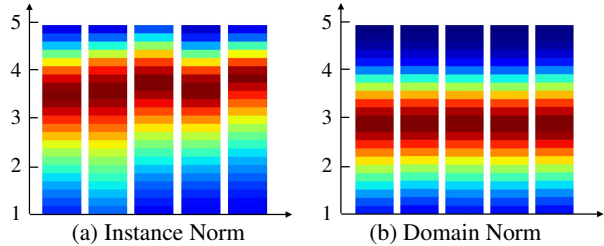

\textbf{Instance normalization (IN)}~\cite{nam2018batch,park2019semantic} overcomes the dependency on data-set statistics by normalizing each sample separately, where elements in $S_i$ are confined to be from the same sample as illustrated in Fig.~\ref{fig:norm}.
In theory, IN is domain-invariant, and normalization across the spatial dimensions ($H$, $W$) reduces image-level appearance/style variations.

However, matching of stereo views is realized at the pixel level by finding an accurate correspondence for each pixel using its $C$-channel feature vector. Any inconsistence of the feature norm and scaling will significantly influence the matching cost and similarity measurements.  

Fig.~\ref{fig:norm2} illustrates that IN cannot regulate the norm distribution of pixel-wise feature vectors that vary in datasets/domains. 

\begin{figure*}[t]
 \setlength{\abovecaptionskip}{14pt}
\setlength{\belowcaptionskip}{-5pt}
 \centering
  \vspace{-2mm}
  \begin{overpic}[width=0.6\linewidth]{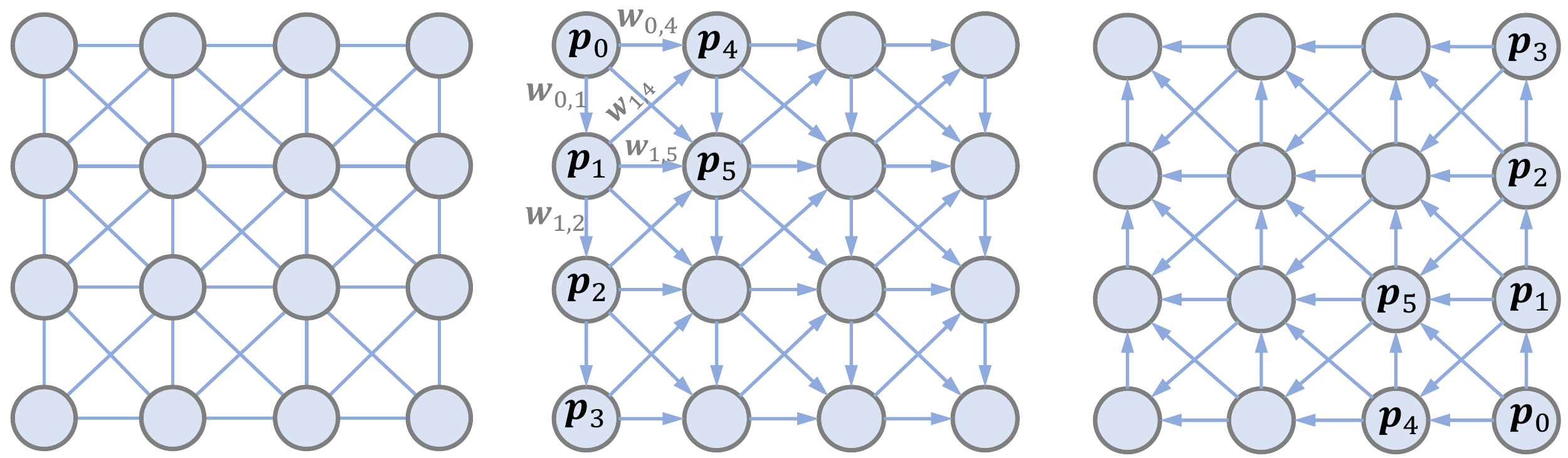}
		\put(2,-2){\small\color{black}{(a) 8-connected graph}}
		\put(37,-2){\small\color{black}{(b) directed graph $G_1$}}
		\put(72,-2){\small\color{black}{(c) directed graph $G_2$}}
	\end{overpic}
 \caption{\small Illustration of the graph construction. The 8-way connected graph is separated into two directed graphs $G_1$ and $G_2$.}
 \label{fig:graph}
 \end{figure*}
 We propose in Fig.~\ref{fig:norm} our \textbf{domain-invariant normalization (DN)}. Our method normalizes features along the spatial axis ($H$, $W$) to induce style-invariant representations similar to IN as well as along the channel dimension ($C$) to enhance the local invariance.

Our DN is realized as follows:
\begin{equation}
\setlength{\abovedisplayskip}{6pt}
\setlength{\belowdisplayskip}{6pt}
 \hat{x}'_{i}=\frac{\hat{x}_i}{\sqrt{\sum_{i\in S'_i}{|\hat{x}_i|^2+\epsilon}}},
 \label{Eq:Norm3}
\end{equation}
where $S'_i$ (green region in Fig.~\ref{fig:norm}) includes $C$ elements from the same example ($N$ axis) and the same spatial location ($H$, $W$ axis).
$\hat{x}_i$ is computed as Eq.~\eqref{Eq:Norm1} and~\eqref{Eq:Norm2} with elements in $S_i$ from the same channel and sample (blue region in Fig.~\ref{fig:norm}).
In DN, besides normalization across spatial dimension,  we also employ $L_2$ normalization to normalize features along the channel axis. They collaborate with each other to address the
address the sensitivity to domain shift  as well as stress noises and extreme values in feature vectors. As illustrated in Fig.~\ref{fig:norm2}, it helps regulate the norm distribution of the features in different datasets
and improves the robustness to local domain shifts (\eg texture pattern, noise, contrast).

Finally, the trainable per-channel scale $\gamma$ and shift $\beta$ are added to enhance the discriminative representation ability as BN and IN. The final formulation is as follows:
\begin{equation}
\setlength{\abovedisplayskip}{6pt}
\setlength{\belowdisplayskip}{6pt}
 y_{i}=\gamma_i~\hat {x}'_i+\beta_i.
 \label{Eq:Norm4}
\end{equation}

\subsection{Non-local Aggregation}\label{sec:agg}

 We propose a graph-based filter that robustly exploits non-local contextual information and reduces the dependence on local patterns (see Fig.~{\color{red}\ref{fig:illustrate}(c)})
 for domain-invariant stereo matching.

\subsubsection{Formulation}
Our inspiration comes from traditional graph-based filters that are remarkably effective in employing non-local structural information for structure-preserving texture and detail removing/smoothing \cite{zhang2015segment}, denoising \cite{zhang2015segment,chen2013fast}, as well as depth-aware estimation and enhancement \cite{liu2013joint,yang2012non}.

For a 2D image/feature map $I$, we construct an 8-connected graph by connecting pixel $\mathbf{p}$ to its eight neighbors (see Fig. \ref{fig:graph}).
To avoid loops and achieve fast non-local information aggregation over the graph,
we split it into two reverse directed graphs $G_1$, $G_2$ (see Fig. \ref{fig:graph}{\color{red}(b)} and \ref{fig:graph}{\color{red}(c)}). 

We assign weight $\mathbf{\omega}_e$ to each edge $e\in G$, and a feature (or color) vector $C(\mathbf{p})$ to each node  $\mathbf{p}\in G$. We also allow $\mathbf{p}$ to propagate information to itself with  weight $\omega_e({\bf p,p})$.
For graph $G_i$ ($i=0,1$), our non-local filter is defined as follows:

\begin{equation}
\setlength{\abovedisplayskip}{0pt}
\setlength{\belowdisplayskip}{2pt}
\begin{array}{rll}
C_i^A(\mathbf{p}) & = & \frac{\sum\limits_{\mathbf{q}\in G_i}{W\mathbf{(q,p)}\cdot C(\mathbf{q})}}{\sum\limits_{\mathbf{q}\in G_i}{W\mathbf{(q,p)}}},\\[4mm]
W\mathbf{(q,p)} &=& \sum\limits_{l_{\mathbf{q},\mathbf{p}}\in G_i}{\prod\limits_{e\in l_{\mathbf{q},\mathbf{p}}}{\mathbf{\omega}_e}}.\\
\end{array}
\label{Eq:filter}
\end{equation}
Here, $l_{\mathbf{q,p}}$ is a feasible path from $\mathbf{q}$ to $\mathbf{p}$.
Note that ${e({\bf q,q})}$ is included in the path and counts for the start node $\mathbf{q}$.
Unlike traditional geodesic filters, we consider all valid paths from source node $\mathbf{q}$ to target node $\mathbf{p}$.
The propagation weight along path $l_{\mathbf{q,p}}$ is the product of all edge weights $\mathbf{\omega}_e$ along the path. Here weight $W\mathbf{(q,p)}$ is defined as the sum of the weights of all feasible paths from $\mathbf{q}$ to $\mathbf{p}$, which determines how much information is diffused to $\mathbf{p}$ from $\mathbf{q}$.

For the edge weight $\omega_{(\mathbf{q,p})}$, we define it in a self-regularized manner as follows:
\begin{equation}
 \setlength{\abovedisplayskip}{6pt}
\setlength{\belowdisplayskip}{8pt}
\begin{array}{ll}
\mathbf{\omega}_e(\mathbf{q,p})=\frac{\mathbf{x_p}^T\mathbf{x_q}}{\|\mathbf{x_p}\|_2\cdot\|\mathbf{x_q}\|_2},
\end{array}
\label{Eq:weights}
\end{equation}
where $\mathbf{x_p}$ and $\mathbf{x_q}$ represent the feature vectors of $\mathbf{p}$ and $\mathbf{q}$, respectively. This definition does not introduce new parameters and thus is more robust to cross-domain generalization.

Compared to other local filters, such as Gaussian filter, median filter, and mean filter that can only propagate information in a local region determined by the filter kernel size, our proposed non-local filter allows the propagation of long-range information with weights as a spatial accumulation along all feasible paths in a graph.

For stable training and to avoid extreme values, we  further add a normalization constraint to the weights associated with $\mathbf{p}$ in the graph $G_i$ as:
\begin{equation}
\setlength{\abovedisplayskip}{6pt}
\setlength{\belowdisplayskip}{6pt}
\sum_{\mathbf{q}\in N_\mathbf{p}}{\mathbf{\omega}_{e(\mathbf{q,p})}} = 1.
\label{Eq:constraint}
\end{equation}
Here, $N_\mathbf{p}$ is the set of the connected neighbors of $\mathbf{p}$ (including itself), and $e(\mathbf{q,p})$ is the directed edge connecting $\mathbf{q}$ and $\mathbf{p}$.
For example, in Fig. {\color{red}\ref{fig:graph}(b)}, for node $\mathbf{p}_0$, $\mathbf{\omega}_{e(\mathbf{p}_0, \mathbf{p}_0)}=1$; and for node $\mathbf{p}_4$, $\mathbf{\omega}_{0,4}+\mathbf{\omega}_{1,4}+\mathbf{\omega}_{e(\mathbf{p}_4, \mathbf{p}_4)}=1$.

If Eq.~\eqref{Eq:constraint} holds, we can further derive $\sum\nolimits_{\mathbf{q}\in G_i}{W\mathbf{(q,p)}}=1$\footnote[1]{\color{red}{The proof is available in the supplementary material.}}.
 Eq.~\eqref{Eq:filter} can then be simplified as follows:
\begin{equation}
\setlength{\abovedisplayskip}{5pt}
\setlength{\belowdisplayskip}{4pt}
\begin{array}{rl}
C_i^A(\mathbf{p})\hspace{2mm}=&\sum\limits_{\mathbf{q}\in G_i}{W\mathbf{(q,p)}\cdot C(\mathbf{q})},
\\[0.4mm]
W\mathbf{(q,p)} =& \sum\limits_{l_{\mathbf{q},\mathbf{p}}\in G_i}{\prod\limits_{e\in l_{\mathbf{q},\mathbf{p}}}{\mathbf{\omega}_e}}.\\
\end{array}
\label{Eq:filter2}
\end{equation}

Such a transformation not only increases the robustness in training but also reduces the computational costs.

\subsubsection{Linear Implementation}
Eq. (\ref{Eq:filter2}) can be realized as an iterative linear aggregation, where the node representation is sequentially updated following the direction of the graph (\eg from top to bottom, then left to right in $G_1$). In each step, $\mathbf{p}$ is updated as:
\begin{equation}
\setlength{\abovedisplayskip}{6pt}
\setlength{\belowdisplayskip}{4pt}
\begin{array}{rl}
C_i^A(\mathbf{p})\hspace{2mm}&=\mathbf{\omega}_{e(\mathbf{p,p})}\cdot{C(\mathbf{p})}+\sum\limits_{\mathbf{q}\in N_\mathbf{p},\mathbf{q}\neq\mathbf{p}}{\mathbf{\omega}_{e\mathbf{(q,p)}}\cdot C_i^A(\mathbf{q})}
\\[3mm]
s.t. &\sum\limits_{\mathbf{q}\in N_\mathbf{p}}{\mathbf{\omega}_{e\mathbf{(q,p)}}=1}.
\end{array}
\label{Eq:linearagg}
\end{equation}

Finally, we repeat the aggregation process for both $G_1$ and $G_2$ where the updated representation with $G_1$ is used as the input for aggregation with $G_2$ (similar to patchmatch stereo \cite{bleyer2011patchmatch}).
The aggregation of Eq.~\eqref{Eq:linearagg} is a linear process with time complexity of $O(n)$ (with $n$ nodes in the graph).
During training, backpropagation can  be realized by reversing the propagation equation which is also a linear process (\textit{available in the supplementary material}).

\subsubsection{Relations to Existing Approaches}
We show that the recently proposed semi-global aggregation (SGA) layer\cite{Zhang2019GANet} and affinity-based propagation approach \cite{liu2017learning} are special cases of our graph-based non-local filter (Eq.~\eqref{Eq:filter2}).  In addition, we compare it with non-local neural networks \cite{wang2018non,xie2019feature} and the attention mechanism \cite{huang2019ccnet}.


{\textbf{Semi-global Aggregation (SGA)}} \cite{Zhang2019GANet} is proposed as a differentiable approximation of SGM \cite{hirschmuller2008stereo} and can be presented as follows:
\begin{equation}
\setlength{\abovedisplayskip}{5pt}
\setlength{\belowdisplayskip}{4pt}
 \begin{array}{rll}
C^A_\mathbf{r}(\mathbf{p},d)&\hspace{-1.7mm}=&\hspace{-4.2mm}\text{sum}\left\{\begin{array}{l}
\hspace{-2mm} {\mathbf{\omega}_0(\mathbf{p},\mathbf{r})} \cdot C(\mathbf{p},d)\\
\hspace{-2mm} {\mathbf{\omega}_1(\mathbf{p},\mathbf{r})}\cdot C^A_\mathbf{r}(\mathbf{p}-\mathbf{r},d)\\
\hspace{-2mm} {\mathbf{\omega}_2(\mathbf{p},\mathbf{r})}\cdot C^A_\mathbf{r}(\mathbf{p}-\mathbf{r},d-1)\\
\hspace{-2mm} {\mathbf{\omega}_3(\mathbf{p},\mathbf{r})}\cdot C^A_\mathbf{r}(\mathbf{p}-\mathbf{r},d+1)\\
\hspace{-2mm} {\mathbf{\omega}_4(\mathbf{p},\mathbf{r})}\cdot \max\limits_{i}{C^A_\mathbf{r}(\mathbf{p}-\mathbf{r},i)}.\\
\end{array}\right.\\[0.9cm]
&s.t.&\sum\limits_{i=0,1,2,3,4}{\mathbf{\omega}_i(\mathbf{p},\mathbf{r})}=1
\end{array}
\label{Eq:sgmlayer2}
\end{equation}

\begin{figure}[t]
 \setlength{\abovecaptionskip}{14pt}
\setlength{\belowcaptionskip}{-3pt}
 \centering
  \vspace{-2mm}
  \hspace{-0.025\linewidth}
  \begin{overpic}[width=1.01\linewidth]{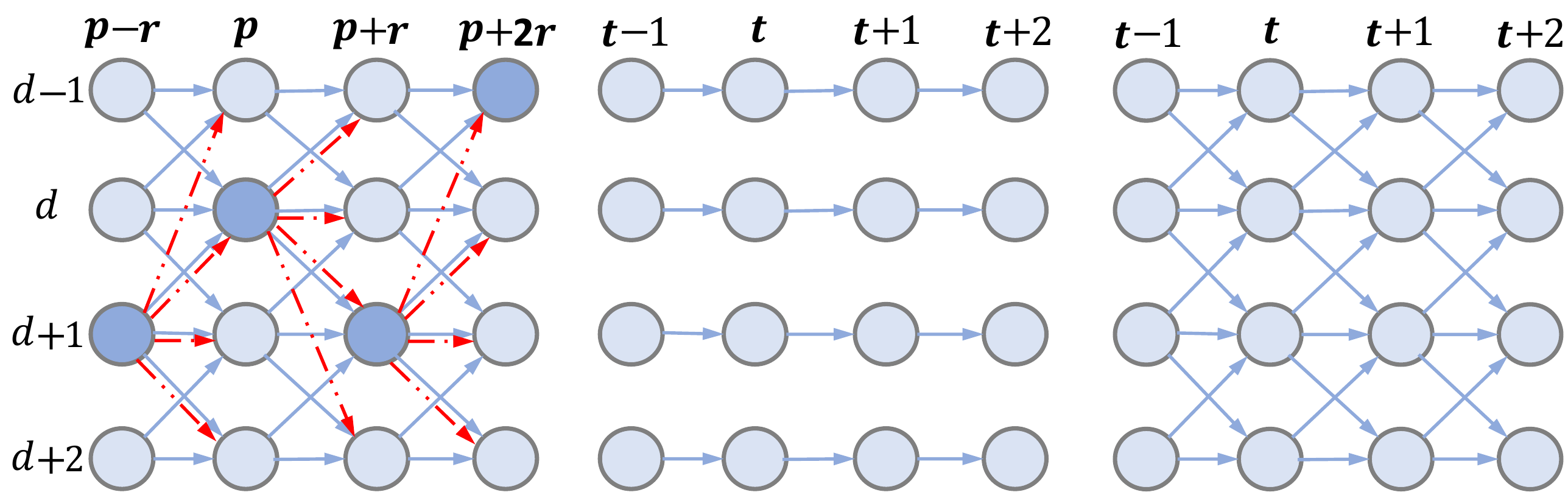}
		\put(6,-3.5){\color{black}{(a) SGA\cite{Zhang2019GANet}}}
		\put(38,-3.5){\color{black}{(b) one-way \cite{liu2017learning}}}
		\put(69.5,-3.5){\color{black}{(c) three-way \cite{liu2017learning}}}
	\end{overpic}
 \caption{\small Special cases of our non-local filter. (a) Semi-global aggregation (SGA) layer \cite{Zhang2019GANet}.  The dark blue node represents the maximum of each column. (b) and (c) are the affinity-based spatial propagations \cite{liu2017learning}. They aggregate from column $t$ to $t+1$.} 
 \label{fig:special}
 \end{figure}

 The aggregations are done in four directions, namely $\mathbf{r}=\{(0,1), (0,-1), (1, 0), (-1, 0)\}$. Taking the right to left propagation ($\mathbf{r}=(0,1)$) as an example, we can construct a propagation graph in Fig. \ref{fig:special}{\color{red}(a)}. The $y$-coordinate represents disparity $d$, and the $x$-coordinate represents the indexes of the pixels/nodes.
 Compared to our non-local graph in Fig. \ref{fig:graph}{\color{red}(b)}, edges connecting top and bottom nodes are removed, and the maximum of each column is densely connected to every node of the next column (red edges).
The SGA layer can then be realized by our proposed non-local filter in Eq.~\eqref{Eq:filter2}.
Here, $(\mathbf{p-r},d\pm 1)$ are the neighborhood nodes of $\mathbf{p}$, and $\mathbf{\omega}_{0,...4}$ are the corresponding edge weights. 

 The {\textbf{Affinity-based Spatial Propagation}} in \cite{liu2017learning} can be achieved as:
\begin{equation}
\setlength{\abovedisplayskip}{5pt}
\setlength{\belowdisplayskip}{8pt}
\begin{array}{lll}
C^A(\mathbf{p},d)&=&\left(1-\sum\limits_{\mathbf{q}\in N_\mathbf{p},\mathbf{q}\neq\mathbf{p}}{\mathbf{\omega}_{e\mathbf{(q,p)}}}\right){C(\mathbf{p})}\\[5mm]
&+&\sum\limits_{\mathbf{q}\in N_\mathbf{p},\mathbf{q}\neq\mathbf{p}}{\mathbf{\omega}_{e\mathbf{(q,p)}} C^A(\mathbf{q})},
\end{array}
\label{Eq:affinity}
\end{equation}
where $\mathbf{\omega}_{e(\mathbf{q,p})}$ are the learned affinities. $1-\sum\nolimits_{\mathbf{q}\in N_\mathbf{p}}{\mathbf{\omega}_{e\mathbf{(q,p)}}}$ is equal to our weight $\mathbf{\omega_{e({\bf p,p})}}$ for $\mathbf{p}$.
The graphs for filtering can be constructed as in Fig. \ref{fig:special}{\color{red}(b)} and \ref{fig:special}{\color{red}(c)} for the one-way and three-way propagations \cite{liu2017learning}, respectively.

\begin{figure*}[t]
 \setlength{\abovecaptionskip}{3pt}
\setlength{\belowcaptionskip}{-10pt}
 \centering
  \vspace{-2mm}
  \hspace{-0.025\linewidth}
  \includegraphics[width=0.8\linewidth,height=0.2\linewidth]{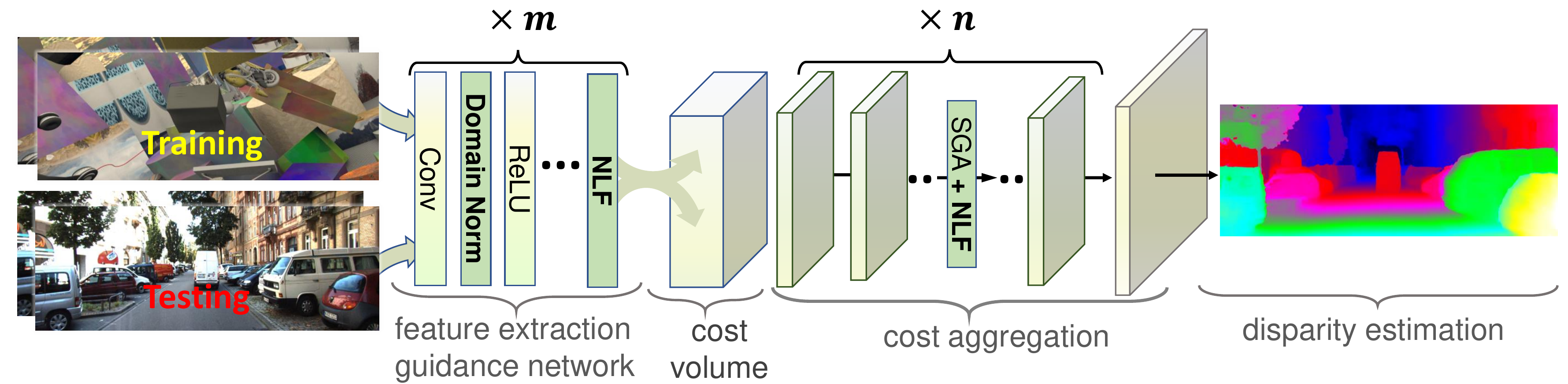}
 \caption{\small Overview of the network architecture. Synthetic data are used for training, while using data from other new domains (\eg real KITTI dataset) for testing. The backbone of the state-of-the-art GANet\cite{Zhang2019GANet} is used as the baseline. The proposed domain normalization is used after each convolutional layer in the feature extraction and guidance network. Several non-local filter layers are implemented for both feature extraction and cost aggregation.}
 \label{fig:network}
 \end{figure*}

The {\textbf{Non-local Neural Networks and Attentions}} \cite{wang2018non,xie2019feature,huang2019ccnet} are implemented without spatial and structural awareness. The similarity definition between two pixels only considers the feature differences without considering their spatial distances.  Therefore, they will easily smooth out depth edges and thin structures (as illustrated in the supplementary material).
 Our non-local filter spatially aggregates the message along the paths in the graph which can avoid over smoothness and better preserve the structure of the disparity maps.

\subsection{Network Architecture}

As illustrated in Fig.~\ref{fig:network}, we utilize the backbone of GANet as the baseline architecture. The local guided aggregation layer in \cite{Zhang2019GANet} is removed since it's domain-dependent and captures a lot of local patterns that are very sensitive to local domain shifts. 

We replace the original batch normalization layer by our proposed domain normalization layer for feature extraction. For the feature extraction network, we utilize a total of seven proposed filtering layers. 
For 3D cost aggregation of the cost volume, two non-local filters are further added for cost volume filtering in each channel/depth. All the details of the network architecture are presented in Table I in the supplementary material.


\section{Experimental Results} \label{sec:experiment}
In our experiments, we train our method only with synthetic data and test it on four real datasets to evaluate its domain generalization ability. 
During training, we use disparity regression \cite{kendall2017end} for disparity prediction, and the smooth $L_1$ loss to compute the errors for back-propagation (the same as in \cite{Zhang2019GANet,chang2018pyramid}).
 All the models are optimized with Adam ($\beta_1 = 0.9$, $\beta_2 = 0.999$).
 We train with a batch size of 8 on four GPUs using $288\times624$
random crops from the input images. The maximum of the disparity is set as 192. We train the model on the synthetic dataset for 10 epochs with a constant learning rate of 0.001. All other training settings are kept the same as those in \cite{Zhang2019GANet}. 

\subsection{Datasets}
{\bf KITTI} stereo 2012 \cite{kitti2012} and 2015 \cite{kitti2015} datasets provide about 400 image pairs of outdoor driving scenes for training, where the disparity labels are transformed from Velodyne LiDAR points. 
The {\bf Cityscapes} dataset \cite{cordts2016cityscapes} provides a large amount of high-resolution ($1k\times2k$) stereo images collected from out-door city driving scenes. The disparity labels are pre-computed by SGM \cite{hirschmuller2008stereo} which is not accurate enough for training deep neural network models.
The {\bf Middlebury} stereo dataset \cite{middleburry}
 is designed for indoor scenes with higher resolution (up to $2k\times 3k$). But it provides no more than 50 image pairs that are not enough to train robust deep neural networks. In addition, {\bf ETH 3D} dataset \cite{ETH3D} provides 27 pairs of gray images for training.

These existing real datasets are all limited by their small quantity or poor ground-truth labels, making them insufficient for training deep learning models.
Hence, we just use them as test sets for evaluating our models' cross-domain generalization ability.

We mainly use synthetic data to train our domain-invariant models. 
The existing Scene Flow synthetic dataset \cite{mayer2016large} contains
35k training image pairs with a resolution of $540\times 960$. 
This dataset has a limited number of the outdoor driving scenes that provide stereo pairs with a few settings of the camera baselines and image resolutions.
We use CARLA \cite{carla} to generate a new supplementary synthetic dataset (with 20k stereo pairs) with more diverse settings, including two kinds of image resolutions ($720\times1080$ and $1080\times1920$), three different focal lengths, and five different camera baselines (in a range of 0.2-1.5m). This supplementary dataset can significantly improve the diversity of the  training set (\textit{which will be published with the paper}).

The two advantages in using synthetic data are that it can avoid all the difficulties of labeling a large amount of real data, and that it can eliminate the negative influence of wrong depth values in real datasets.

\subsection{Ablation Study}
We evaluate the performance of our DSMNet with numerous settings, including different architectures, normalization strategies and numbers (0-9) of the proposed non-local filter (NLF) layers. As listed in Table \ref{tab:ablation}, the full-setting DSMNet far outperforms the baseline in accuracy by 3\% on the KITTI and 8\% on the Middlebury datasets. Our proposed domain normalization improves the accuracy by about 1.5\%, and the NLF layers contribute another 1.4\% on the KITTI dataset.


\begin{table}[t]
	\setlength{\abovecaptionskip}{5pt}
	\setlength{\belowcaptionskip}{-2pt}
	\vspace{-2mm}
	\centering
	\footnotesize
	\caption{\small Ablation study. Models are trained on synthetic data (SceneFlow). Threshold error rates (\%) are used for evaluations.}
	\begin{tabular}{C{0.9cm}| C{1cm} C{1.2cm}| C{1cm}| C{0.8cm} C{0.8cm}}
	\hline
      \multirow{2}[0]{*}{\tabincell{c}{\hspace{-1mm}Normlize}} &\multicolumn{2}{c|}{Non-local Filter}&\multirow{2}[0]{*}{\tabincell{c}{\hspace{-1mm}Backbone}}&{Midd} & \hspace{-1mm}KITTI \\
     & feature &\hspace{-2mm}cost volume& &3-pixel&2-pixel  \\
	\hline\hline
     BN& &  &ours& 30.3&9.4\\
     DN& &  &ours& 27.1&7.9\\
     DN& +3 &  &ours &24.2&7.1\\
     DN& +7 &  &ours &22.9&6.8\\
     DN& +9 &  &ours &22.4&6.8\\
     DN& +7 & +2  &ours &{\bf 21.8}&{\bf 6.5}\\
     \hline\hline
     BN&    &     &PSMNet&39.5 & 16.3\\
     BN&    &     &GANet&32.2 &11.7\\
     DN& +7 & +2  &PSMNet& 26.1& 8.5\\
     DN& +7 & +2  &GANet&23.7 &7.3 \\

   \hline
  \end{tabular}
\label{tab:ablation}%
\vspace*{-0.075in}
\end{table}

Moreover, our proposed layers are generic and could be seamlessly integrated into other deep stereo matching models. Here, we replace our backbone model with GANet \cite{Zhang2019GANet} and PSMNet \cite{chang2018pyramid}. The accuracies are improved by 4$\sim$8\% on KIITTI dataset and 8$\sim$13\% on Middlebury dataset for coss-domain evaluations compared with the original PSMNet and GANet.

\begin{figure*}[t]
 \setlength{\abovecaptionskip}{-1pt}
\setlength{\belowcaptionskip}{-10pt}
 \centering
 \hspace{-2mm}
 \subfigure[Input view]{
  \begin{minipage}[b]{0.247\linewidth}
    \includegraphics[width=1\linewidth,height=0.56\linewidth]{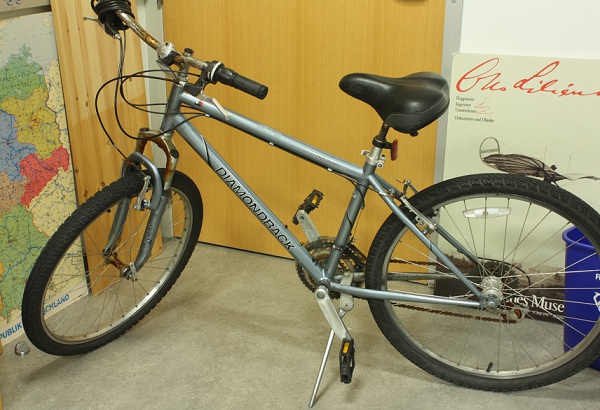}\\
    \includegraphics[width=1\linewidth]{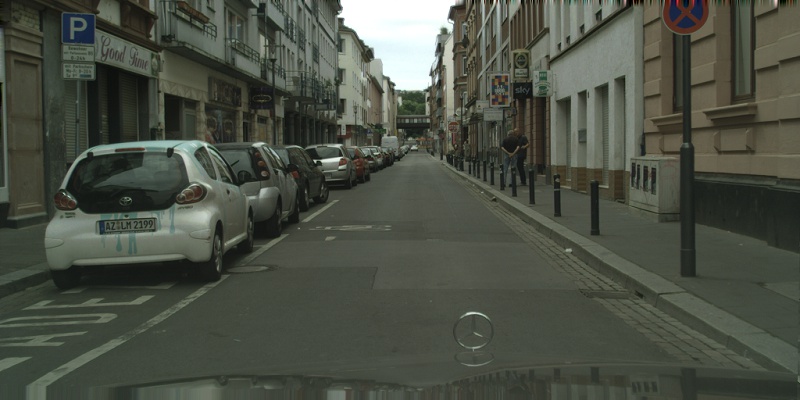}
  \end{minipage}
}
\hspace{-3mm}
 \subfigure[HD$^3$\cite{yin2019hierarchical}]{
  \begin{minipage}[b]{0.247\linewidth}
    \includegraphics[width=1\linewidth,height=0.56\linewidth]{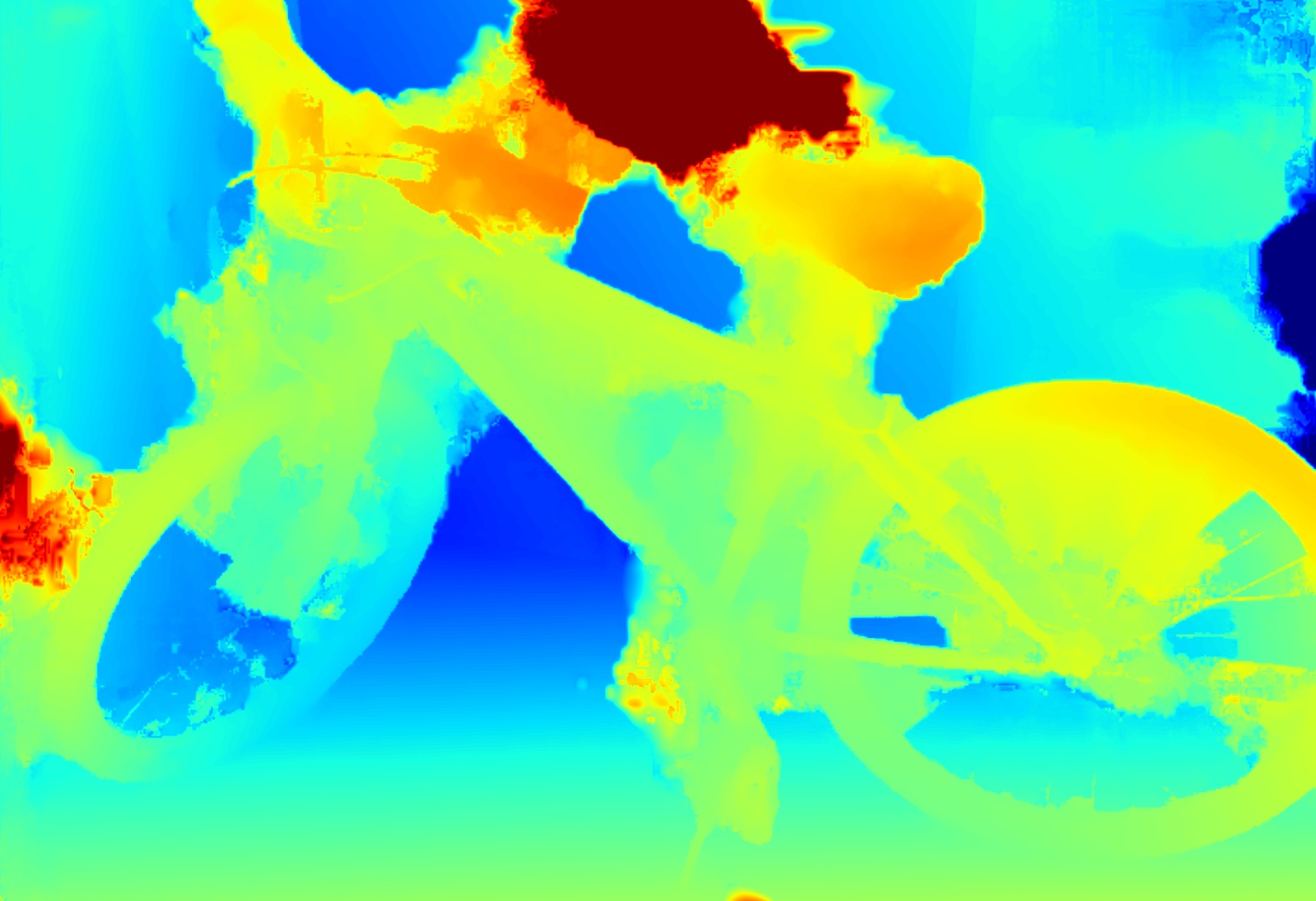}\\
    \includegraphics[width=1\linewidth]{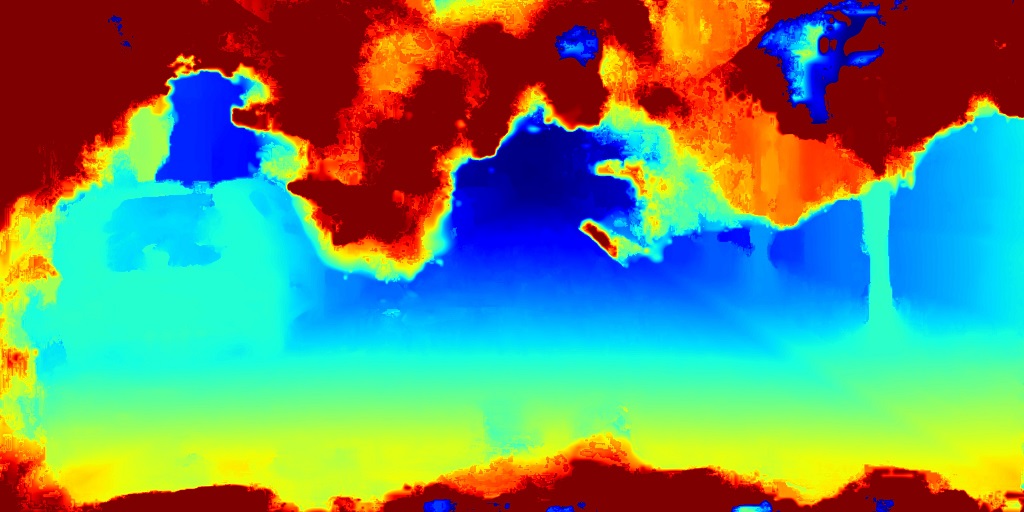}
  \end{minipage}
	}
\hspace{-3mm}
 \subfigure[PSMNet\cite{chang2018pyramid}]{
   \begin{minipage}[b]{0.247\linewidth}
  \includegraphics[width=1\linewidth,height=0.56\linewidth]{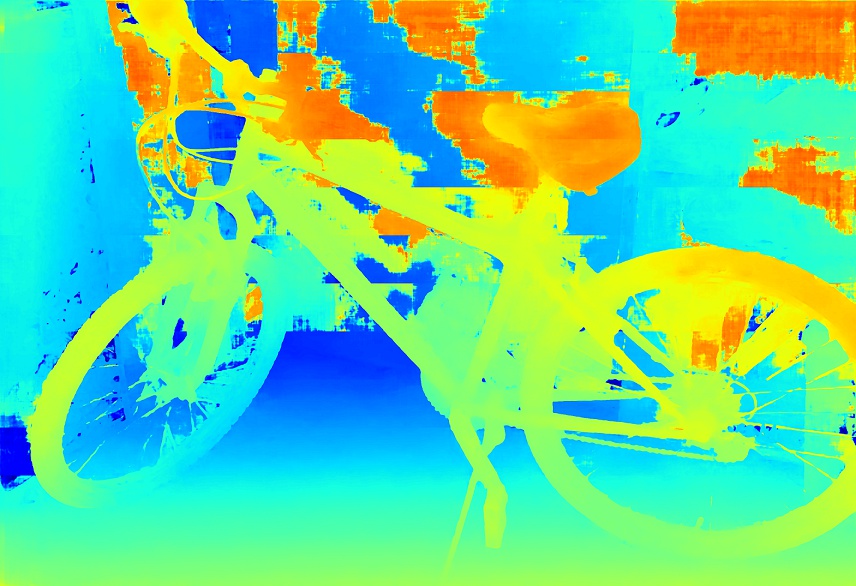}\\
    \includegraphics[width=1\linewidth]{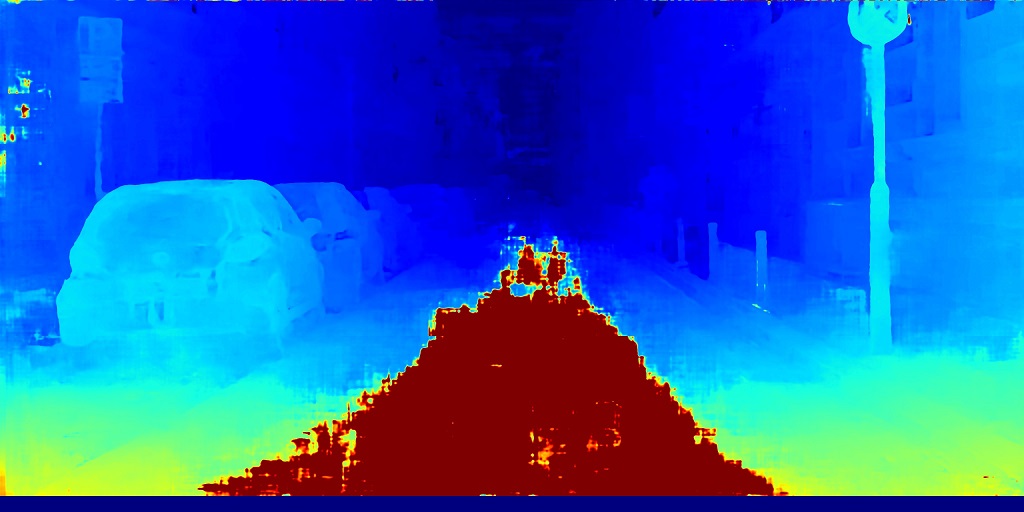}
  \end{minipage}
}
\hspace{-3mm}
 \subfigure[Our DSMNet]{
  \begin{minipage}[b]{0.247\linewidth}
  \includegraphics[width=1\linewidth,height=0.56\linewidth]{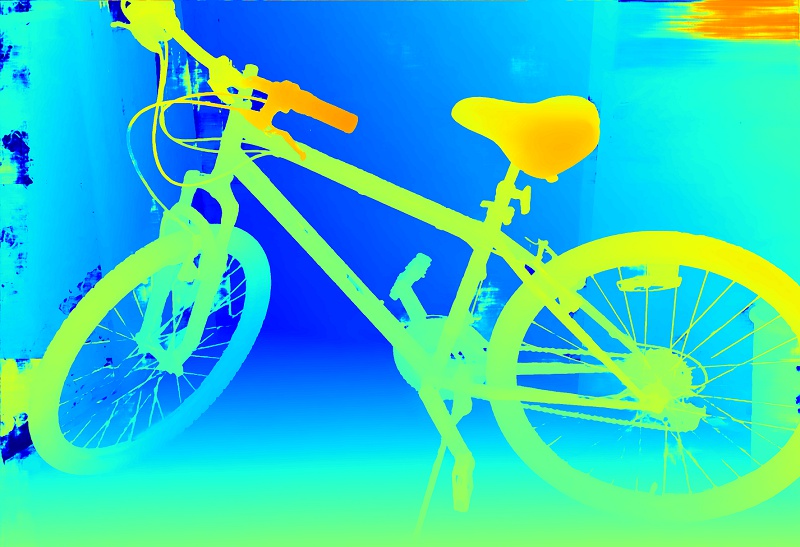}\\
    \includegraphics[width=1\linewidth]{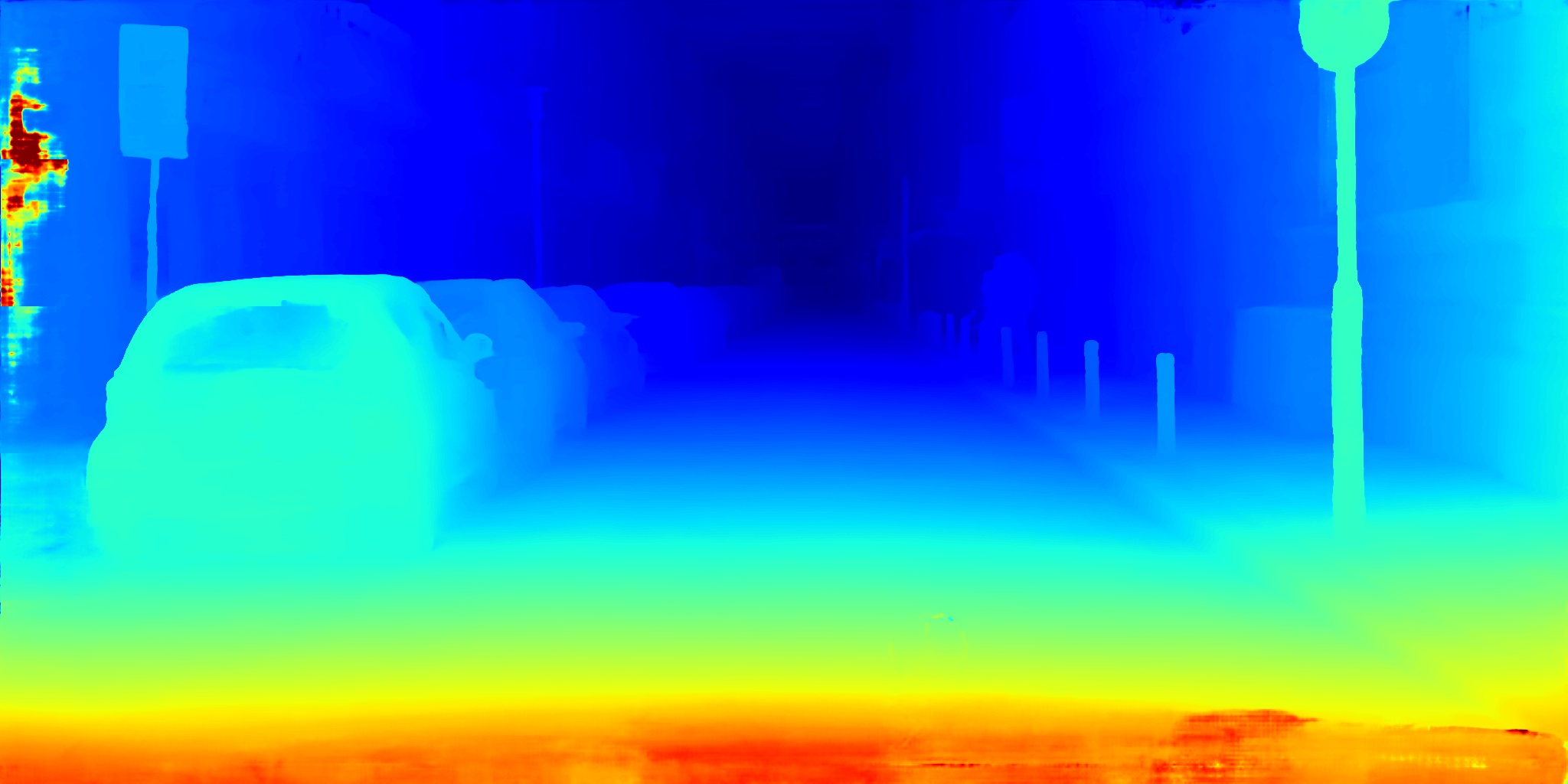}
  \end{minipage}
 }
 \caption{\small Comparisons with state-of-the-art models. Models are trained on synthetic data and evaluated on high-resolution real datasets (Middlebury and CityScapes). Our DSMNet can produce much more accurate disparity estimation. (See supplementary for more results.} 
 \label{fig:results}
 \end{figure*}

\subsection{Component Analysis and Comparisons}
To further validate the superiorities of the proposed layers , we compare each of them with other related normalization and non-local strategies.
\begin{table}[t]
\setlength{\abovecaptionskip}{5pt}
\setlength{\belowcaptionskip}{0pt}
\centering
\footnotesize
\caption{\small Comparisons with Existing Normalization and Filtering/Attention Strategies}
\begin{tabular}{R{3cm} C{2cm} C{1.5cm}}
\hline
Models   & Middlebury (full) & KITTI\\
 \hline\hline
 Batch Norm & 29.1 & 7.3 \\
 Instance Norm   &27.1 &6.4 \\
 Adaptive Norm\cite{nam2018batch}  & 28.2 & 6.8 \\
 Attention\cite{huang2019ccnet}  & 25.2 & 5.9 \\
 Feature Denoising\cite{xie2019feature}  &25.9 &6.1 \\
 Affinity \cite{liu2017learning} & 23.1 & 5.2 \\
 DSMNet (full setting) & 20.1 & 4.1 \\
 \hline
\end{tabular}
\label{tab:norm_filter}%
\vspace*{-0.075in}
\end{table}
\vspace{-4mm}
\paragraph{Normalization Strategies.}
Table \ref{tab:norm_filter} compares our domain normalization with batch normalization \cite{ioffe2015batch}, instance normalization \cite{ulyanov2016instance}, and the recently proposed adaptive batch-instance normalization \cite{nam2018batch}. We keep all other settings the same as our DSMNet and only replace the normalization method for training and evaluation. Our domain normalization is superior to others for domain-invariant stereo matching because it can fully regulate the feature vectors' distribution and remove both image-level and local contrast differences for cross-domain generalization. 
\vspace{-4mm}
\paragraph{Non-local Approaches.}
 Finally, we compare our graph-based non-local filter with other related strategies, including affinity-based propagation \cite{liu2017learning}, non-local neural network denoising \cite{xie2019feature}, and non-local attention \cite{huang2019ccnet} (in Table \ref{tab:norm_filter}). Our graph-based filtering strategy is better for capturing the structural and geometric context for robust domain-invariant stereo matching. The non-local neural network denoising \cite{xie2019feature} and non-local attention \cite{huang2019ccnet} do not have spatial constraints that usually lead to over smoothness of the depth edges (as shown in the supplementary material). Affinity-based propagations \cite{liu2017learning} are special cases of our proposed filtering strategy and are not as effective in feature and cost volume aggregations for stereo matching.

\begin{table}[t]
	\setlength{\abovecaptionskip}{5pt}
	\setlength{\belowcaptionskip}{0pt}
	\centering
	\footnotesize
	\caption{\small Evaluations on the KITTI, Middlebury, and ETH 3D validation datasets. Threshold error rates (\%) are  used.} 
	\begin{tabular}{R{1.45cm}|| L{0.35cm} L{0.35cm} | L{0.4cm} L{0.4cm} L{0.4cm}| C{0.705cm}| C{0.475cm}}
	\hline

    Models& \multicolumn{2}{c|}{\hspace{-1mm}\tabincell{c}{KITTI\\\hspace{-0mm}2012~~2015}}
    &\multicolumn{3}{c|}{\tabincell{c}{Middlebury\\~full\quad~half~~~quarter}}&{\hspace{-1.95mm}\tabincell{c}{ETH3D}}&\hspace{-2mm}\tabincell{c}{Carla}\\

	\hline\hline
  \hspace{-2mm}CostFilter\cite{hosni2013fast}&21.7 &18.9 &57.2 & 40.5& 17.6 & 31.1&41.1 \\
  \hspace{-3mm}PatchMatch\cite{bleyer2011patchmatch}&20.1 &17.2 &50.2 & 38.6 & 16.1 &24.1&30.1 \\ 
  SGM\cite{hirschmuller2008stereo}&7.1 &7.6 &38.1 &25.2 &10.7 & 12.9&20.2 \\ 
  \hline
  Training set& \multicolumn{7}{c}{SceneFlow}\\
  \hline
  HD$^3$\cite{yin2019hierarchical}&23.6 & 26.5 & 50.3 & 37.9 & 20.3 & 54.2&35.7\\
  gwcnet\cite{guo2019group}& 20.2& 22.7 &47.1 &34.2 & 18.1 & 30.1&33.2\\
  PSMNet\cite{chang2018pyramid}&15.1 &16.3 &39.5 &25.1 &14.2 &23.8&25.9\\
  GANet\cite{Zhang2019GANet}&10.1 &11.7 &32.2 & 20.3& 11.2&14.1&18.8\\
  \hspace{-2mm}{\bf Our DSMNet}&{\bf 6.2} &{\bf 6.5}&{\bf 21.8}&{\bf 13.8}&{\bf 8.1}&{\bf 6.2} &{\bf 9.8}\\
  \hline
  Training set& \multicolumn{7}{c}{SceneFlow + Carla}\\
  \hline
  HD$^3$\cite{yin2019hierarchical}&19.1 & 19.5 & 47.3 & 35.2 & 19.5 & 45.2&--\\
  gwcnet\cite{guo2019group}& 17.2& 18.1 &45.2 &31.8 & 17.2 & 29.4&--\\
  PSMNet\cite{chang2018pyramid}&10.3 &11.0 &35.5 &23.7 &13.8 &20.3&--\\
  GANet\cite{Zhang2019GANet}&7.2 &7.6 &31.9 & 19.7& 11.4&13.5&--\\
  \hspace{-2mm}{\bf Our DSMNet}&{\bf 3.9}&{\bf 4.1}&{\bf20.1} &{\bf 13.6}&{\bf 8.2}&{\bf 6.0}&--\\
   \hline
  \end{tabular}
\label{tab:comparison}%
\vspace*{-0.05in}
\end{table}
\subsection{Cross-Domain Evaluations}
In this section, we compare our proposed DSMNet with state-of-the-art stereo matching models by training with synthetic data and evaluating on  real test sets. 
\vspace{-4mm}
\paragraph{Comparisons with State-of-the-Art Models.}
In Table \ref{tab:comparison} and Fig. \ref{fig:results}, we compare our DSMNet with other state-of-the-art deep neural network models on the four real datasets. All the models are trained on  synthetic data (either SceneFlow or a mixture of SceneFlow and Carla). We find that DSMNet far outperforms the state-of-the-art models by 3$\sim$30\% in error rates on all these datasets. It is also far better than traditional stereo matching algorithms, like SGM \cite{hirschmuller2008stereo}, costfilter \cite{hosni2013fast} and patchmatch \cite{bleyer2011patchmatch}.
\vspace{-4mm}

\begin{table}[t]
\setlength{\abovecaptionskip}{5pt}
\setlength{\belowcaptionskip}{0pt}
\centering
\footnotesize
\caption{\small Evaluation on KITTI 2015 Benchmark}
\begin{tabular}{C{2.8cm}  C{2cm} C{2cm}}
\hline
Models~~~~~~&Training Set& Error Rates (\%)\\
 \hline\hline
 {\bf Our DSMNet}&{\bf Synthetic} & {\bf 3.71} \\
 MC-CNN-acrt\cite{zbontar2015computing}& Kitti-gt & 3.89 \\
 DispNetC\cite{mayer2016large}  & Kitti-gt & 4.34 \\
 Content-CNN\cite{luo2016efficient}  & Kitti-gt & 4.54 \\
 MADNet-finetune\cite{tonioni2019real}  & Kitti-gt & 4.66 \\
 Weak Supervise\cite{tulyakov2017weakly} & Kitti-gt& 4.97 \\
  MADNet\cite{tonioni2019real}  &Kitti (no gt) & 8.23 \\
 OASM-Net\cite{li2018occlusion}&Kitti (no gt)&	8.98\\
 Unsupervised\cite{zhou2017unsupervised} &Kitti (no gt)& 9.91\\
 \hline
\end{tabular}
\label{tab:kitti2015}%
\vspace*{-0.05in}
\end{table}
 \begin{figure*}[t]
 \setlength{\abovecaptionskip}{-1pt}
\setlength{\belowcaptionskip}{-5pt}
 \centering
  \vspace{-2mm}
 \hspace{-2mm}
 \subfigure[Input view]{
    \includegraphics[width=0.33\linewidth]{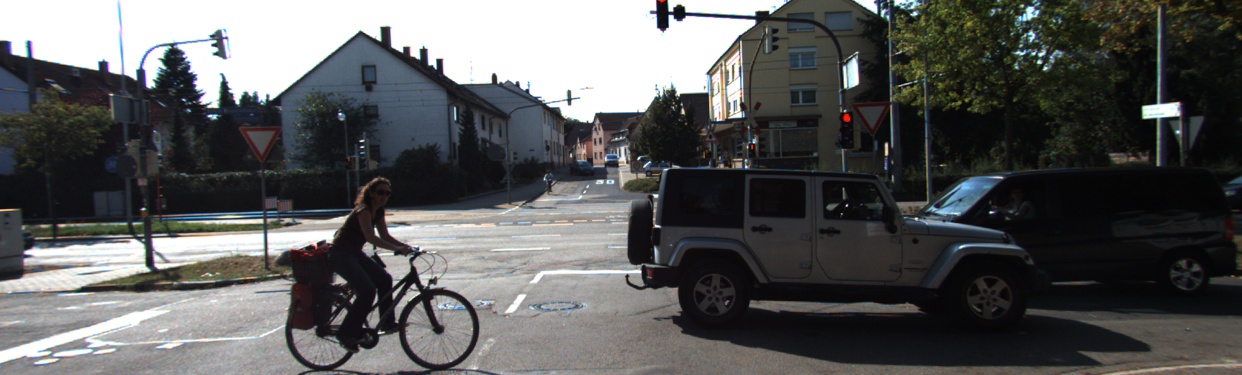}
    }
  \hspace{-3mm}
 \subfigure[MC-CNN \cite{zbontar2015computing}]{
    \includegraphics[width=0.33\linewidth]{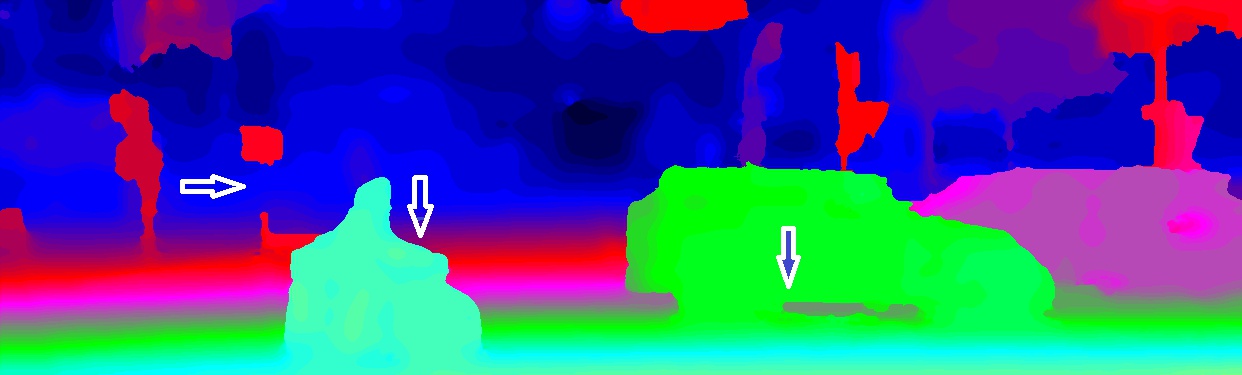}
    }
  \hspace{-3mm}
 \subfigure[PSMNet \cite{chang2018pyramid}]{
    \includegraphics[width=0.33\linewidth]{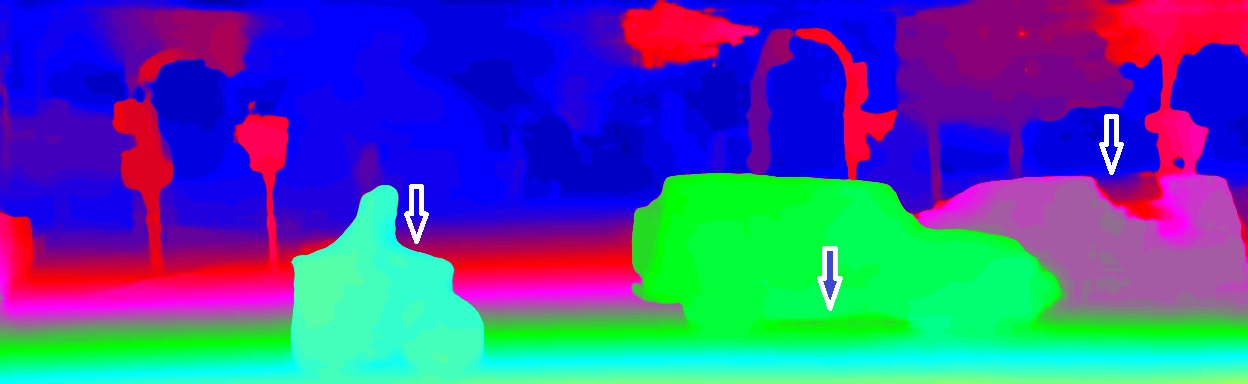}
    }
  \\[-3mm]
  \hspace{-2mm}
 \subfigure[HD$^3$ \cite{yin2019hierarchical}]{
    \includegraphics[width=0.33\linewidth]{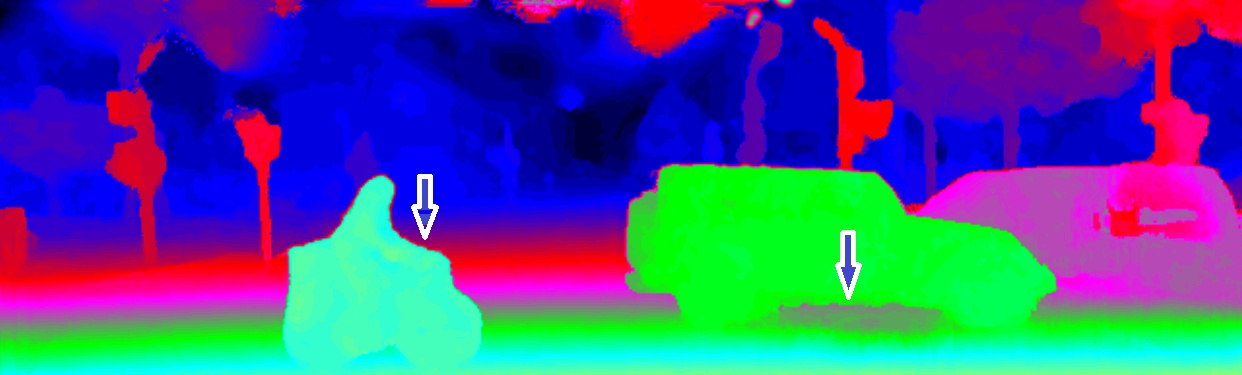}
    }
  \hspace{-3mm}
 \subfigure[GANet-deep \cite{Zhang2019GANet}]{
    \includegraphics[width=0.33\linewidth]{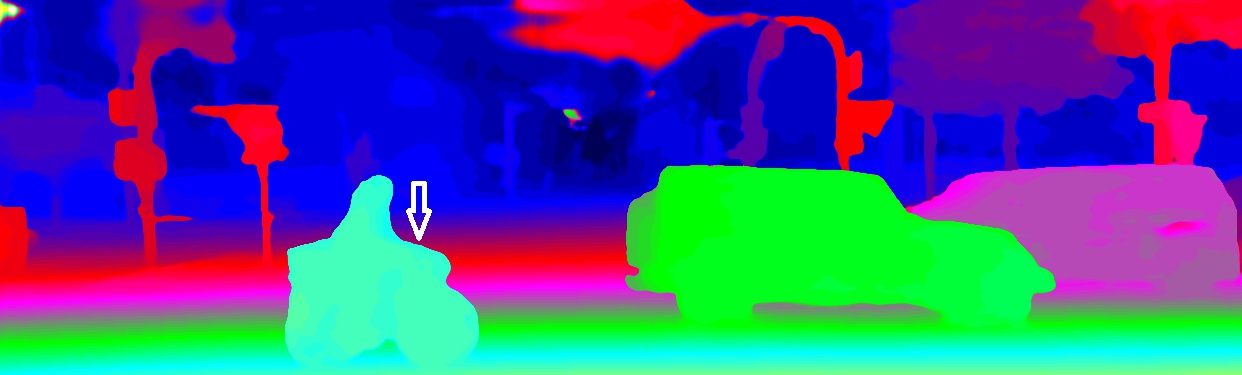}
    }
 \hspace{-3mm}
 \subfigure[Our DSMNet-synthetic]{
    \includegraphics[width=0.33\linewidth]{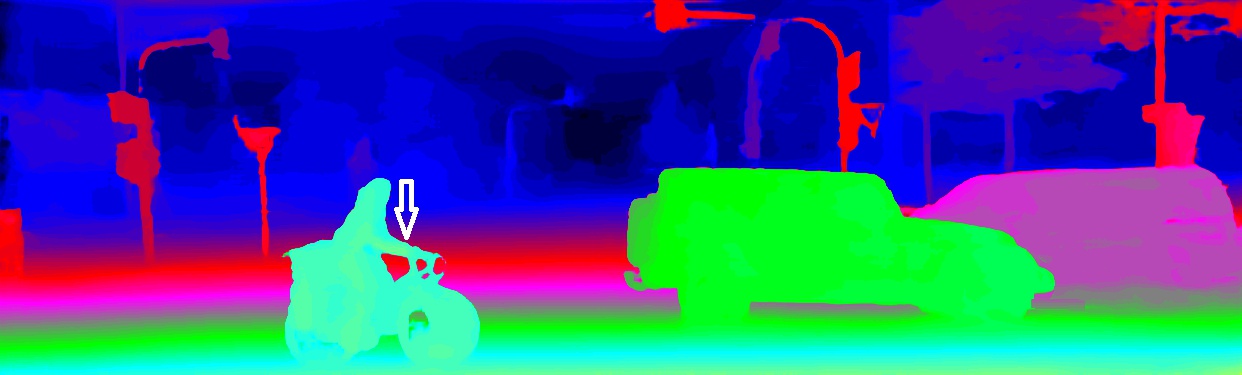}
    }
 \caption{\small Comparisons with the fine-tuned state-of-the-art models. Our model is trained only with synthetic data. All others are fine-tuned on the KITTI target scenes. As pointed by  arrows,  our DSMNet can produce more accurate object boundaries.}%
 \label{fig:finetune}
 \end{figure*}
\paragraph{Evaluation on the KITTI Benchmark.}
Table \ref{tab:kitti2015} presents the performance of our DSMNet  on the KITTI benchmark \cite{kitti2015}. Our model far outperforms most of the unsupervised/self-supervised models trained on the KITTI domain. It is even better than supervised stereo matching networks (including, MC-CNN\cite{zbontar2015computing}, content-CNN\cite{luo2016efficient}, and DispNetC \cite{mayer2016large}) trained or fine-tuned on the KITTI dataset. When compared with other fine-tuned state-of-the-art models (\eg PSMNet\cite{chang2018pyramid}, HD$^3$\cite{yin2019hierarchical}, GANet-deep\cite{Zhang2019GANet}), our DSMNet (without fine-tuning) produces more accurate object boundaries (Fig.~\ref{fig:finetune}).

\subsection{Fine-tuning}
In this section, we show DSMNet's best performance when fine-tuned on the target domain.
We fine-tune the model pre-trained on synthetic data for a further 700 epochs using the KITTI 2015 training set. The learning rate for fine-tuning begins at 0.001 for the first 300 epochs and decreases to 0.0001 for the rest. The results are submitted to the KITTI benchmarks for evaluations.

Table \ref{tab:kitti2015_finetune} compares the results of the fine-tuned DSMNet and those of other state-of-the-art DNN models. 
We find that DSMNet outperforms most of the recent models (including PSMNet \cite{chang2018pyramid}, HD$^3$ \cite{yin2019hierarchical}, GwcNet \cite{guo2019group} and GANet-15\cite{Zhang2019GANet}) by a noteworthy margin. This implies that DSMNet can achieve the same accuracy by fine-tuning on one specific dataset, without sacrificing accuracy to improve its cross-domain generalization ability.

We also separately test the effectiveness of our non-local filtering strategy. Using the current best ``GANet-deep''\cite{Zhang2019GANet} (including the Local Guided Aggregation layer) as the baseline, we add five filtering layers for feature extraction.  All other settings are kept the same as the original GANet. After training on synthetic data and fine-tuning on the KITTI training dataset, the  model gets a new state-of-the-art accuracy (1.77\%) on KITTI 2015 benchmark. 
This shows that our graph-based filter can improve not only cross-domain generalization but also the accuracy on the test domains.
\begin{table}[t]
	\setlength{\abovecaptionskip}{3pt}
	\setlength{\belowcaptionskip}{0pt}
	\centering
	\footnotesize
	\caption{\small Evaluation on the KITTI 2015 Benchmark (Fine-tuning)}
	\begin{tabular}{C{3cm} C{2cm} C{2cm}}
		\hline
		Models  & Non-Occluded & All Area\\
		\hline\hline
		GANet + {\bf Our NLF} &{\bf 1.58} &  {\bf 1.77}\\
		GANet-deep\cite{Zhang2019GANet}&1.63 &1.81  \\
		{\bf DSMNet-finetune}& 1.71 & 1.90  \\
		GANet-15\cite{Zhang2019GANet}&1.73  &1.93  \\
		HD$^3$\cite{yin2019hierarchical}&1.87  &2.02 \\
		gwcnet-g\cite{guo2019group} &1.92  &2.11 \\
		PSMNet\cite{chang2018pyramid}& 2.14   & 2.32 \\
		GCNet\cite{kendall2017end} &2.61 &2.87 \\
		\hline
	\end{tabular}
	\label{tab:kitti2015_finetune}%
	\vspace*{-0.05in}
\end{table}
\subsection{Efficiency and Parameters}
Our proposed non-local filtering is a linear process that can be realized efficiently. The inference time is increased slightly by no more than 5\% compared with the baseline. 
Moreover, no any new parameter is introduced for the proposed domain normalization and non-local filtering layers. 
\textit{Detailed comparisons are available in the supplementary material.}

\section{Conclusion}
In this paper, we have proposed two end-to-end trainable neural network layers for our domain-invariant stereo matching network. Our novel domain normalization can fully regulate the distribution of learned features to address significant domain shifts, and our non-local graph-based filter can capture more robust non-local structural and geometric features for accurate disparity estimation in cross-domain situations.
 We have verified our model on four real datasets and have shown its superior accuracy when compared to other state-of-the-art stereo matching networks in cross-domain generalization.

\newpage
{\small
	\bibliographystyle{ieee_fullname}
	\bibliography{mybib}
}
\newpage
\onecolumn
\appendix{\centering{\section*{\Large Supplementary Material}}}

\section{Proof of Footnote {\color{red}1}}

Following all the variable definitions in the paper, here, we prove that
\begin{equation}
\sum\limits_{\mathbf{q}\in G_i}{W\mathbf{(q,p)}}=1,~~~\text{if}~~\sum\limits_{\mathbf{q}\in N_\mathbf{p}}{\mathbf{\omega}_{e(\mathbf{q,p})}}=1.
\label{Eq:conclusion}
\end{equation}

Since any path which reaches node $\mathbf{p}$ must pass through its neighborhoods $\mathbf{q}$, we can expand $W(\mathbf{q,p})$ to get that $$\sum\limits_{\mathbf{q}\in G_i}{W\mathbf{(q,p)}}={\mathbf \omega}_{e(\mathbf{p, p})}+\sum\limits_{\mathbf{p'}\in N_\mathbf{p}, \mathbf{p'}\neq \mathbf{p}}{\mathbf{\omega}_{e(\mathbf{ p',p})}\sum\limits_{\mathbf{q}\in G_i}{W(\mathbf{q, p'})}}$$

Following the order of $\mathbf{p}_0, \mathbf{p}_1...\mathbf{p}_n...\mathbf{p}_N$ (Fig.~\ref{fig:graph}), we can prove Eq.~\eqref{Eq:conclusion} by mathematical induction:

When $n=0$, for $\mathbf{p}_0$, $\sum\limits_{\mathbf{q}\in G_i}{W(\mathbf{q},\mathbf{p}_0)}=W(\mathbf{p}_0,\mathbf{p}_0)=\omega_{e(\mathbf{p}_0,\mathbf{p}_0)}=1$

Assume when $n\le t$, $\sum\limits_{\mathbf{q}\in G_i}{W(\mathbf{q},\mathbf{p}_n)}=1$.

We can get that for $n = t+1$:
$$\begin{array}{lll}
\sum\limits_{\mathbf{q}\in G_i}W(\mathbf{q}, \mathbf{p}_{t+1})&=&{\mathbf \omega}_{e(\mathbf{p}_{t+1}, \mathbf{p}_{t+1})}+\sum\limits_{\mathbf{p}_{k}\in N_{\mathbf{p}_{t+1}}, \mathbf{p}_{k}\neq \mathbf{p}_{t+1}}{\mathbf{\omega}_{e(\mathbf{p}_k,\mathbf{p}_{t+1})}\sum\limits_{\mathbf{q}\in G_i}{W(\mathbf{q},\mathbf{p}_k)}}\\[5mm]
&=&\mathbf{ \omega}_{e(\mathbf{p}_{t+1}, \mathbf{p}_{t+1})}+\sum\limits_{\mathbf{p}_{k}\in N_{\mathbf{p}_{t+1}}, \mathbf{p}_{k}\neq \mathbf{p}_{t+1}}{\mathbf{\omega}_{e(\mathbf{p}_{k},\mathbf{p}_{t+1})}\cdot 1}\\[5mm]
&=&\sum\limits_{\mathbf{p}_{k}\in N_{\mathbf{p}_{t+1}}}{\mathbf{\omega}_{e(\mathbf{p}_{k},\mathbf{p}_{t+1})}}\\[5mm]
&=&1.
\end{array}
$$

Here, $k\le t$, since $\mathbf{p}_{k}\in N_{\mathbf{p}_{t+1}}$.

This yields the equivalence of Eq.~\eqref{Eq:conclusion}.

\section{Backpropagation}

The backpropagation for $\mathbf{\omega}_e$ and $C(\mathbf{p})$ in Eq.~\eqref{Eq:linearagg} can be computed inversely. Assume the gradient from next layer is $\frac{\partial E}{\partial C^A_i}$. The backpropagation can be implemented as: 
\begin{equation}
\begin{array}{l}
\frac{\partial E}{\partial C(\mathbf{p})}={\frac{\partial E}{\partial C^b_{i}(\mathbf{p})}\cdot \mathbf{\omega}_{e(\mathbf{p},\mathbf{p})}},\\[4mm]
\frac{\partial E}{\partial \mathbf{\omega}_{e(\mathbf{p},\mathbf{p})}}={\frac{\partial E}{\partial C^b_{i}(\mathbf{p})}\cdot C(\mathbf{p})},\\[4mm]
\frac{\partial E}{\partial \mathbf{\omega}_{e(\mathbf{q},\mathbf{p})}}={\frac{\partial E}{\partial C^b_{i}(\mathbf{p})}\cdot C^A_i(\mathbf{q})},~~\mathbf{q}\in N_{\mathbf{p}}~\&~\mathbf{q}\neq\mathbf{p}
\end{array}
\end{equation}
where, $\frac{\partial E}{\partial C^b_{\mathbf{i}}}$ is a temporary gradient variable which can be calculated iteratively (similar to Eq.~\eqref{Eq:linearagg}):
\begin{equation}
\frac{\partial E}{\partial C^b_{i}(\mathbf{p})} = \frac{\partial E}{\partial C^A_{i}(\mathbf{p})}+\sum\limits_{\mathbf{q}\in N_\mathbf{p},\mathbf{q}\neq\mathbf{p}}{\frac{\partial E}{\partial C^b_i(\mathbf{q})} \cdot \mathbf{\omega}_e(\mathbf{q},\mathbf{p})}
\label{EQ:backnlf}
\end{equation}

The propagation of Eq.~\eqref{EQ:backnlf} is an inverse process and in an order of $\mathbf{p}_N, \mathbf{p}_{N-1},...\mathbf{p}_0$

\section{Details of the Architecture}
Table \ref{tab:architecture} presents the details of the parameters of the DSMNet.  It has seven non-local filtering layers which are used in feature extraction and cost aggregation. The proposed Domain Normalization layer is used to replace Batch Normalization after each 2D convolutional layer in the feature extraction and guidance networks.

\section{Efficiency and Parameters}
As shown in Table \ref{tab:time}, our proposed non-local filtering is a linear process that can be realized efficiently. The inference time is increased by about 5\% compared with the baseline. 
Moreover, no any new parameters are introduced for the proposed domain normalization and non-local filtering layers. 

\iftrue
\begin{table}[h]
	\setlength{\abovecaptionskip}{5pt}
	\setlength{\belowcaptionskip}{0pt}
	\centering
	\footnotesize
	\caption{\small Efficiency (Elapsed Time) and Number of Parameter}
	\begin{tabular}{R{3cm}  C{3cm} C{3cm}}
		\hline
		{Methods}  & {Elapsed Time} &Parameter Number\\
		\hline\hline
		GANet-deep \cite{Zhang2019GANet} & 1.8s& 60M\\
		Baseline & {1.4s} &  48M\\
		Our DSMNet & 1.5s& 48M\\
		PSMNet \cite{chang2018pyramid} & 0.4s &52M\\
		DSMNet (PSMNet) & 0.42s &52M\\
		\hline
	\end{tabular}
	\label{tab:time}%
	\vspace*{-0.075in}
\end{table}
\fi

\section{Carla Dataset}
Since the synthetic Sceneflow dataset \cite{mayer2016large} only has limited number £¨about 7,000£© of stereo pairs for diving scenes, we use the Carla \cite{carla} platform to produce the stereo pairs for outdoor driving scenes. As shown in Table \ref{tab:carladata}, the new carla supplementary dataset has more diverse settings, including two kinds of image resolutions ($720\times1080$ and $1080\times1920$), three different focal lengths, and six different camera baselines (in a range of 0.2-1.5m). This supplementary dataset can significantly improve the diversity of the training set. As shown in Fig.~\ref{fig:carlaexample}, the Carla scenes still have significant domain differences (\eg color, textures) compared with the real scenes (\eg KITTI, CityScapes), but, our DSMNet can extract shape and structure information for robust stereo matching. These can be better transferred to the real scenes and produce more accurate disparity estimation.
\iftrue
\begin{table*}[t]
	\setlength{\abovecaptionskip}{5pt}
	\setlength{\belowcaptionskip}{0pt}
	\centering
	\footnotesize

	\caption{\small Statistics of the Carla Stereo Dataset}
	\begin{tabular}{C{2.5cm} C{2.5cm}  C{3cm} C{3cm} C{3cm}}
		\hline
		dataset&{number of pairs}  & {focal length} &baseline settings & resolutions\\
		\hline\hline
		SceneFlow &34,000& 450, 1050 & 0.54& $960\times540$\\
		Carla Stereo &20,000& 640, 670, 720 & 0.2, 0.3, 0.5, 1.0, 1.2, 1.5& $1280\times720$, $1920\times1080$\\
		\hline
	\end{tabular}
	\label{tab:carladata}%
\end{table*}
\fi
\begin{table*}[t]
	\renewcommand\arraystretch{0.95}
	\setlength{\abovecaptionskip}{5pt}
	\setlength{\belowcaptionskip}{10pt}
	\centering
	\footnotesize
	\vspace{-2mm}
	\caption{\small Parameters of the network architecture of ``DSMNet''}
	\begin{tabular}{C{2cm}|L{6cm} |L{4cm}}
		\hline
		{\bf No.} & {\bf Layer Description} & {\bf Output Tensor}\\
		\hline\hline
		\multicolumn{3}{c}{\bf Feature Extraction} \\
		\hline
		input& normalized image pair as input & H$\times$W$\times$3 \\
		1 & 3$\times$3 conv, {\bf DN}, ReLU & H$\times$W$\times$32\\
		2 & 3$\times$3 conv, stride 3, {\bf DN}, ReLU  & $\sfrac{1}{3}$H$\times$$\sfrac{1}{3}$W$\times$32\\
		3 & 3$\times$3 conv, {\bf DN}, ReLU& $\sfrac{1}{3}$H$\times$$\sfrac{1}{3}$W$\times$32\\
		4 & {\bf NLF}, {\bf DN}, ReLU& $\sfrac{1}{3}$H$\times$$\sfrac{1}{3}$W$\times$32\\
		5 & 3$\times$3 conv, stride 2, {\bf DN}, ReLU   & $\sfrac{1}{6}$H$\times$$\sfrac{1}{6}$W$\times$48\\
		6 & {\bf NLF}, {\bf DN}, ReLU& $\sfrac{1}{6}$H$\times$$\sfrac{1}{6}$W$\times$48\\
		7 & 3$\times$3 conv, {\bf DN}, ReLU   & $\sfrac{1}{6}$H$\times$$\sfrac{1}{6}$W$\times$48\\
		8-9 & repeat 5,7 & $\sfrac{1}{12}$H$\times$$\sfrac{1}{12}$W$\times$64\\
		10-11 & repeat 8-9 & $\sfrac{1}{24}$H$\times$$\sfrac{1}{24}$W$\times$96\\
		12-13 & repeat 8-9& $\sfrac{1}{48}$H$\times$$\sfrac{1}{48}$W$\times$128\\
		14 & 3$\times$3 deconv, stride 2, {\bf DN}, ReLU  & $\sfrac{1}{24}$H$\times$$\sfrac{1}{24}$W$\times$96\\
		15 & 3$\times$3 conv, {\bf DN}, ReLU  & $\sfrac{1}{24}$H$\times$$\sfrac{1}{24}$W$\times$96\\
		16-17 & repeat 14-15 & $\sfrac{1}{12}$H$\times$$\sfrac{1}{12}$W$\times$64\\
		18-19 & repeat 14-15 & $\sfrac{1}{6}$H$\times$$\sfrac{1}{6}$W$\times$48\\
		20 & {\bf NLF}, {\bf DN}, ReLU& $\sfrac{1}{6}$H$\times$$\sfrac{1}{6}$W$\times$48\\
		21-22 & repeat 14-15 & $\sfrac{1}{3}$H$\times$$\sfrac{1}{3}$W$\times$32\\
		23-41 & repeat 4-22& $\sfrac{1}{3}$H$\times$$\sfrac{1}{3}$W$\times$32\\
		42 & {\bf NLF}, {\bf DN}, ReLU& $\sfrac{1}{3}$H$\times$$\sfrac{1}{3}$W$\times$32\\
		\hline
		concatenation& 	\multicolumn{2}{l}{(11,14) (9,16) (7,18) (4,21) (20,24) (17,27) (15,29) (13,31) (18,25) (30,33) (28,35) (26,37) (23, 40)} \\
		\hline
		{\tabincell{c}{\hspace{-1mm}cost volume}}& by feature concatenation & $\sfrac{1}{3}$H$\times$$\sfrac{1}{3}$W$\times$64$\times$32\\
		\hline
		\multicolumn{3}{c}{\bf Guidance Branch} \\
		\hline
		input& concate 1 and up-sampled 35 as input& H$\times$W$\times$64\\
		(1)&  3$\times$3 conv, {\bf DN}, ReLU  & H$\times$W$\times$16\\
		(2) & 3$\times$3 conv, stride 3, {\bf DN}, ReLU  & $\sfrac{1}{3}$H$\times$$\sfrac{1}{3}$W$\times$32\\
		(3) & 3$\times$3 conv, {\bf DN}, ReLU  & $\sfrac{1}{3}$H$\times$$\sfrac{1}{3}$W$\times$32\\
		(4) & 3$\times$3 conv (no bn \& relu)& $\sfrac{1}{3}$H$\times$$\sfrac{1}{3}$W$\times$20\\
		(5) &  split, reshape, normalize & $4\times$ $\sfrac{1}{3}$H$\times$$\sfrac{1}{3}$W$\times$5\\
		(6)-(8) &  from (3), repeat (3)-(5) & $4\times$ $\sfrac{1}{3}$H$\times$$\sfrac{1}{3}$W$\times$5\\
		(9)-(11) & from (6), repeat (6)-(8) &$4\times$ $\sfrac{1}{3}$H$\times$$\sfrac{1}{3}$W$\times$5\\
		(12) & from (2), 3$\times$3 conv, stride 2, {\bf DN}, ReLU  & $\sfrac{1}{6}$H$\times$$\sfrac{1}{6}$W$\times$32\\
		(13) & 3$\times$3 conv, {\bf DN}, ReLU  & $\sfrac{1}{6}$H$\times$$\sfrac{1}{6}$W$\times$32\\
		(14) & 3$\times$3 conv (no bn \& relu)& $\sfrac{1}{6}$H$\times$$\sfrac{1}{6}$W$\times$20\\
		(15) &  split, reshape, normalize & $4\times$ $\sfrac{1}{6}$H$\times$$\sfrac{1}{6}$W$\times$5\\
		(16)-(18) & from (13), repeat (13)-(15) &$4\times$ $\sfrac{1}{6}$H$\times$$\sfrac{1}{6}$W$\times$5\\
		(19)-(21) & from (16), repeat (13)-(15) &$4\times$ $\sfrac{1}{6}$H$\times$$\sfrac{1}{6}$W$\times$5\\
		(22)-(24) & from (19), repeat (13)-(15) &$4\times$ $\sfrac{1}{6}$H$\times$$\sfrac{1}{6}$W$\times$5\\

		\hline
		\multicolumn{3}{c}{\bf Cost Aggregation} \\
		\hline
		input& 4D cost volume & $\sfrac{1}{3}$H$\times$$\sfrac{1}{3}$W$\times$64$\times$64\\
		$[1]$& 3$\times$3$\times$3, 3D conv & $\sfrac{1}{3}$H$\times$$\sfrac{1}{3}$W$\times$64$\times$32\\
		$[2]$&  { SGA}: weight matrices from (5) &$\sfrac{1}{3}$H$\times$$\sfrac{1}{3}$W$\times$64$\times$32\\
		$[3]$&  {\bf NLF} &$\sfrac{1}{3}$H$\times$$\sfrac{1}{3}$W$\times$64$\times$32\\
		$[4]$& 3$\times$3$\times$3, 3D conv & $\sfrac{1}{3}$H$\times$$\sfrac{1}{3}$W$\times$64$\times$32\\
		\multirow{2}[0]{*}{\tabincell{c}{output}} &3$\times$3$\times$3, 3D to 2D conv, upsamping & H$\times$W$\times$193\\
		& softmax, regression, loss weight: 0.2& H$\times$W$\times$1\\
		$[5]$& 3$\times$3$\times$3, 3D conv, stride 2 & $\sfrac{1}{6}$H$\times$$\sfrac{1}{6}$W$\times$32$\times$48\\
		$[6]$& 3$\times$3$\times$3, 3D conv & $\sfrac{1}{6}$H$\times$$\sfrac{1}{6}$W$\times$32$\times$48\\
		$[7]$&  { SGA}: weight matrices from (15) &$\sfrac{1}{6}$H$\times$$\sfrac{1}{6}$W$\times$32$\times$48\\
		$[8]$& 3$\times$3$\times$3, 3D conv, stride 2 & $\sfrac{1}{12}$H$\times$$\sfrac{1}{12}$W$\times$16$\times$64\\
		$[9]$& 3$\times$3$\times$3, 3D deconv, stride 2 & $\sfrac{1}{6}$H$\times$$\sfrac{1}{6}$W$\times$32$\times$48\\
		$[10]$& 3$\times$3$\times$3, 3D conv & $\sfrac{1}{6}$H$\times$$\sfrac{1}{6}$W$\times$32$\times$48\\
		$[11]$&  { SGA}: weight matrices from (18) &$\sfrac{1}{6}$H$\times$$\sfrac{1}{6}$W$\times$32$\times$48\\
		$[12]$& 3$\times$3$\times$3, 3D deconv, stride 2 & $\sfrac{1}{3}$H$\times$$\sfrac{1}{3}$W$\times$64$\times$32\\
		$[13]$& 3$\times$3$\times$3, 3D conv & $\sfrac{1}{3}$H$\times$$\sfrac{1}{3}$W$\times$64$\times$32\\
		$[14]$& { SGA}: weight matrices from (8) &$\sfrac{1}{3}$H$\times$$\sfrac{1}{3}$W$\times$64$\times$32\\
		$[15]$&  {\bf NLF} &$\sfrac{1}{3}$H$\times$$\sfrac{1}{3}$W$\times$64$\times$32\\
		\multirow{2}[0]{*}{\tabincell{c}{output}} &3$\times$3$\times$3, 3D to 2D conv, upsamping & H$\times$W$\times$193\\
		& softmax, regression, loss weight: 0.6 & H$\times$W$\times$1\\
		$[16-26]$ & repeat $[5-15]$ &$\sfrac{1}{3}$H$\times$$\sfrac{1}{3}$W$\times$64$\times$32\\
		\multirow{2}[0]{*}{\tabincell{c}{final\\output}} &3$\times$3$\times$3, 3D to 2D conv, upsamping & H$\times$W$\times$193\\
		& regression, loss weight: 1.0 & H$\times$W$\times$1\\

		\hline
		{connection} & \multicolumn{2}{l}{concate: (4,12), (7,9), (8,19), (11,16), (15,23), (18,20); add: (1,4)} \\
		\hline
	\end{tabular}
	\label{tab:architecture}%
	\vspace*{-0.075in}
\end{table*}
\begin{figure*}[h]
	\setlength{\abovecaptionskip}{-3pt}
	\setlength{\belowcaptionskip}{-10pt}
	\centering
	\vspace{-2mm}
	\subfigure[left view]{
		\includegraphics[width=0.32\linewidth]{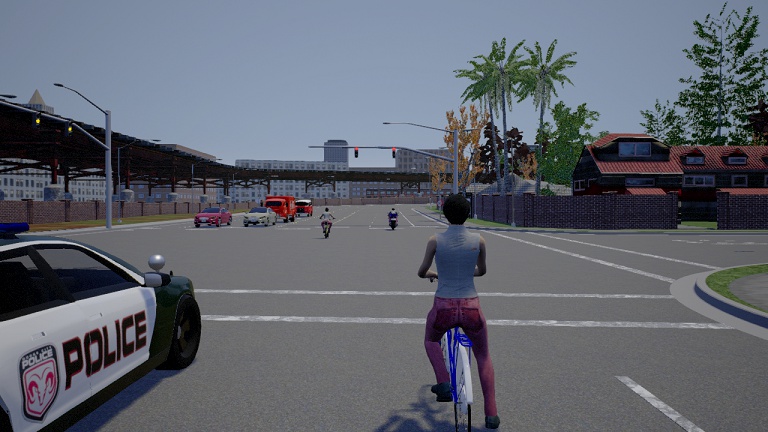}
	}
	\hspace{-2mm}
	\subfigure[right view]{
		\includegraphics[width=0.32\linewidth]{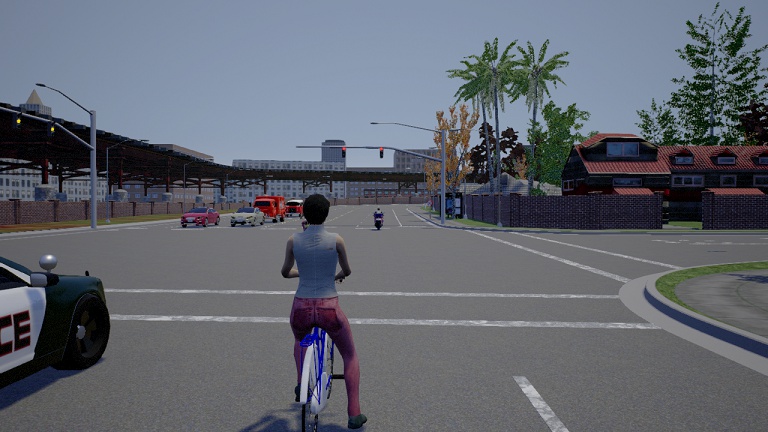}
	}
	\hspace{-2mm}
	\subfigure[disparity map]{
		\includegraphics[width=0.32\linewidth]{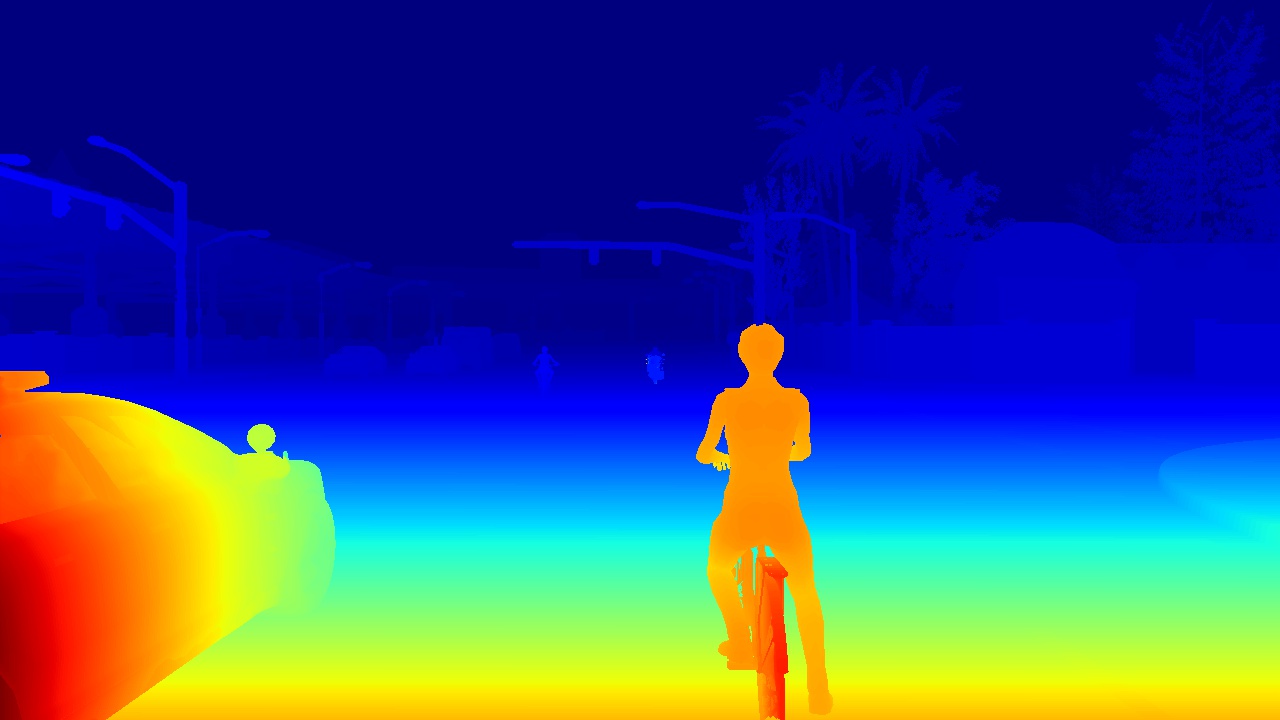}
	}
	\caption{\small Example of the Carla stereo data.}
	\label{fig:carlaexample}
\end{figure*}


\section{More Results}
\subsection{Feature Visualization}
\begin{figure*}[t]
	\setlength{\abovecaptionskip}{-3pt}
	\setlength{\belowcaptionskip}{-10pt}
	\centering
	\vspace{-2mm}
	\hspace{-2mm}
	\subfigure[Input view]{
		\begin{minipage}[b]{0.32\linewidth}
			\includegraphics[width=1\linewidth,height=0.28\linewidth]{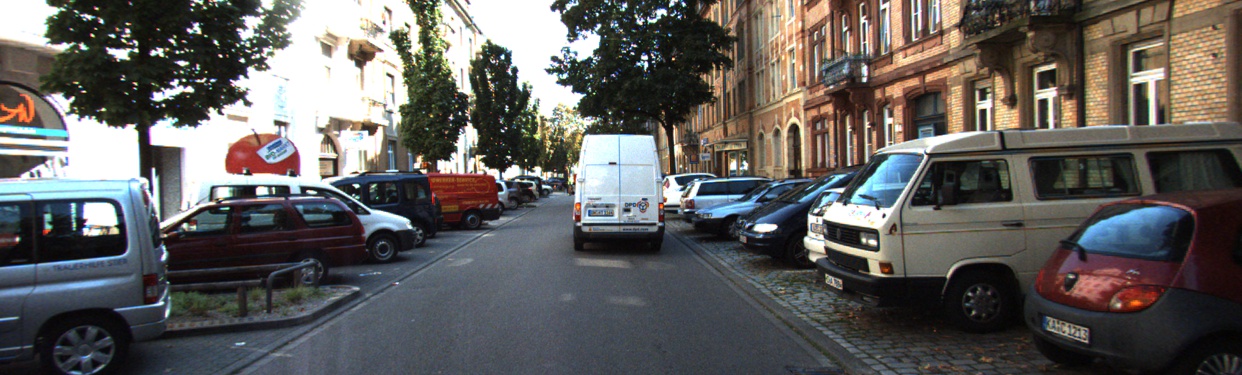}\\
			\includegraphics[width=1\linewidth,height=0.65\linewidth]{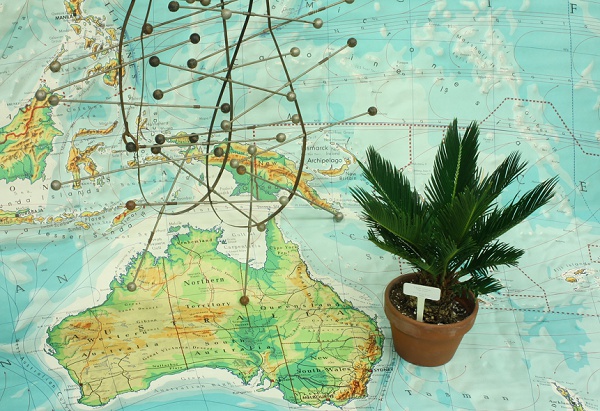}\\
			\includegraphics[width=1\linewidth,height=0.5\linewidth]{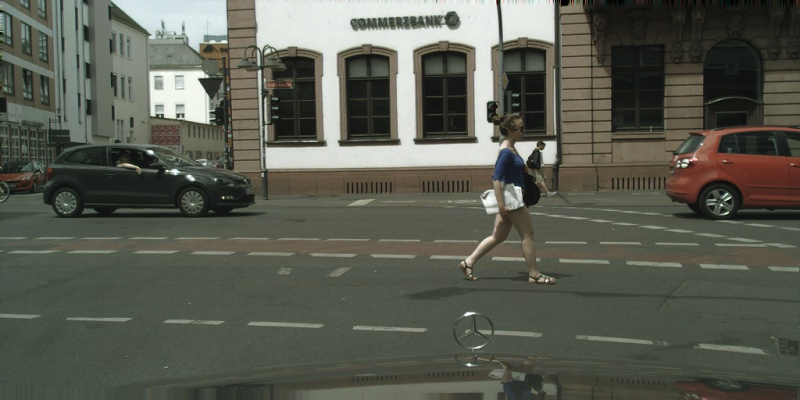}
		\end{minipage}
	}
	\hspace{-3mm}
	\subfigure[GANet-synthetic]{
		\begin{minipage}[b]{0.32\linewidth}
			\includegraphics[width=1\linewidth,height=0.28\linewidth]{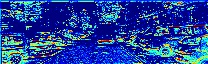}\\
			\includegraphics[width=1\linewidth,height=0.65\linewidth]{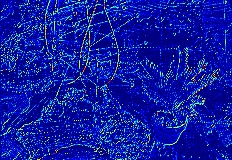}\\
			\includegraphics[width=1\linewidth,height=0.5\linewidth]{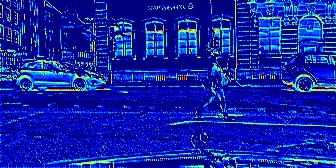}
		\end{minipage}
	}
	\hspace{-3mm}
	\subfigure[GANet-finetune]{
		\begin{minipage}[b]{0.32\linewidth}
			\includegraphics[width=1\linewidth,height=0.28\linewidth]{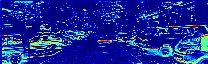}\\
			\includegraphics[width=1\linewidth,height=0.65\linewidth]{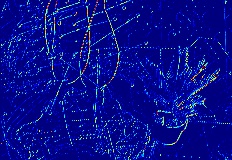}\\
			\includegraphics[width=1\linewidth,height=0.5\linewidth]{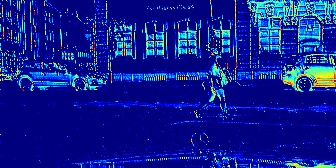}
		\end{minipage}
	}\\[-2mm]
	\subfigure[HD$^3$-synthetic]{
		\begin{minipage}[b]{0.32\linewidth}
			\includegraphics[width=1\linewidth,height=0.28\linewidth]{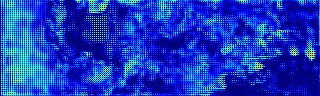}\\
			\includegraphics[width=1\linewidth,height=0.65\linewidth]{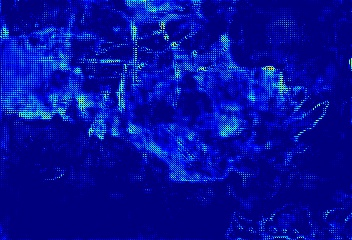}\\
			\includegraphics[width=1\linewidth,height=0.5\linewidth]{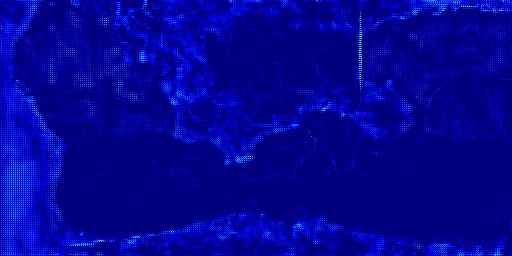}
		\end{minipage}
	}
	\hspace{-3mm}
	\subfigure[PSMNet-synthetic]{
		\begin{minipage}[b]{0.32\linewidth}
			\includegraphics[width=1\linewidth,height=0.28\linewidth]{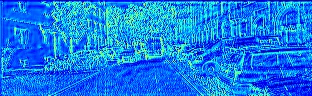}\\
			\includegraphics[width=1\linewidth,height=0.65\linewidth]{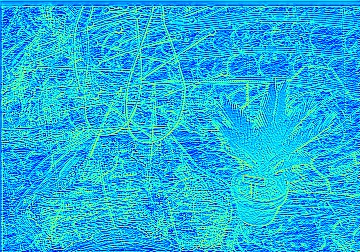}\\
			\includegraphics[width=1\linewidth,height=0.5\linewidth]{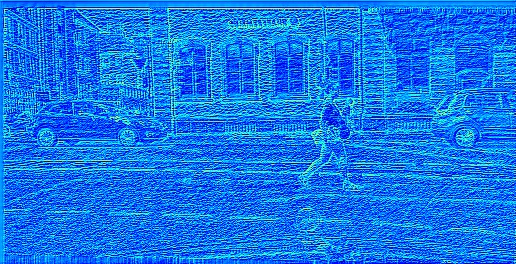}
		\end{minipage}
	}
	\hspace{-3mm}
	\subfigure[DSMNet-synthetic]{
		\begin{minipage}[b]{0.32\linewidth}
			\includegraphics[width=1\linewidth,height=0.28\linewidth]{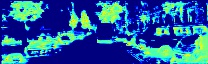}\\
			\includegraphics[width=1\linewidth,height=0.65\linewidth]{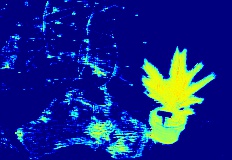}\\
			\includegraphics[width=1\linewidth,height=0.5\linewidth]{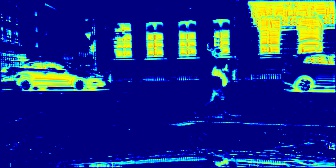}
		\end{minipage}
	}
	\caption{\small Comparison and visualization of the feature maps for cross-domain test . (b) GANet \cite{Zhang2019GANet}, (d) HD$^3$ \cite{yin2019hierarchical}, (e) PSMNet \cite{chang2018pyramid} are trained on the synthetic dataset (Sceneflow \cite{mayer2016large}) and test on other real scenes/datasets (from top to bottom: Kitti \cite{kitti2015}, Middlebury \cite{middleburry} and CityScapes \cite{cordts2016cityscapes}). The features are mainly local patterns and produce a lot of artifacts (\eg noises) when suffering from domain shifts. (c) GANet is finetuned on the test dataset for comparisons. The artifacts have been stressed after fine tuning. (f) Our DSMNet trained on the synthetic data. No distortions and artifacts are introduced on the feature maps. It mainly captures the non-local structure and shape information, which are more robust for cross-domain generalization.}
	\label{fig:featuremore}
\end{figure*}

As compared in Fig.~\ref{fig:featuremore}, the features of the state-of-the-art models are mainly local patterns which can have a lot of artifacts (\eg noises) when suffering from domain shifts. Our DSMNet  mainly captures the non-local structure and shape information, which are robust for cross-domain generalization. There is no artifacts in the feature maps of our DSMNet.

\subsection{Disparity Results on Different Datasets}
More results and comparisons are provided in Fig.~\ref{fig:resultsmore}. All the models are trained on the synthetic dataset and tested on the real KITTI, Middlebury, ETH3D and Cityscapes datasets.

\iftrue
\begin{figure*}[t]
	\setlength{\abovecaptionskip}{-3pt}
	\setlength{\belowcaptionskip}{-10pt}
	\centering
	\vspace{-2mm}
	\hspace{-2mm}
	\subfigure[Input view]{
		\begin{minipage}[b]{0.247\linewidth}
			\includegraphics[width=1\linewidth]{images/input_5.jpg}\\
			\includegraphics[width=1\linewidth]{images/input_42.jpg}\\
			\includegraphics[width=1\linewidth,height=0.65\linewidth]{images/middlebury.jpg}\\
			\includegraphics[width=1\linewidth,height=0.65\linewidth]{images/middlebury2.jpg}\\
			\includegraphics[width=1\linewidth]{images/cityscape2.jpg}\\
			\includegraphics[width=1\linewidth]{images/cityscape22.jpg}
			\includegraphics[width=1\linewidth,height=0.55\linewidth]{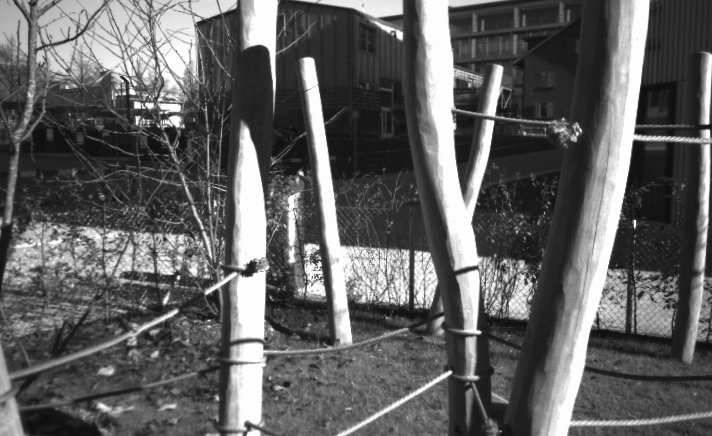}\\
			\includegraphics[width=1\linewidth,height=0.55\linewidth]{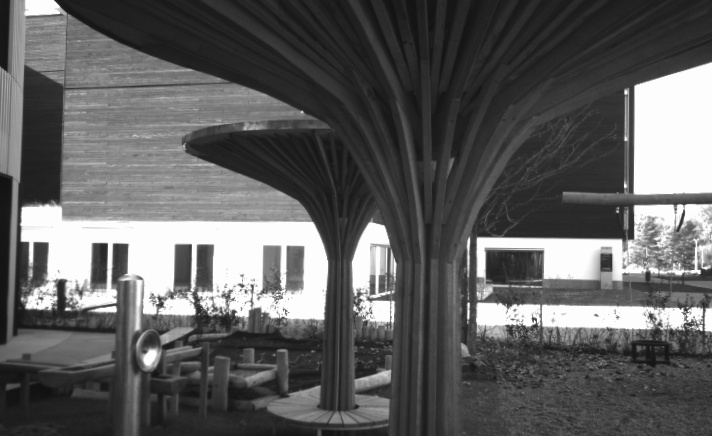}
		\end{minipage}
	}
	\hspace{-3mm}
	\subfigure[HD$^3$\cite{yin2019hierarchical}]{
		\begin{minipage}[b]{0.247\linewidth}
			\includegraphics[width=1\linewidth]{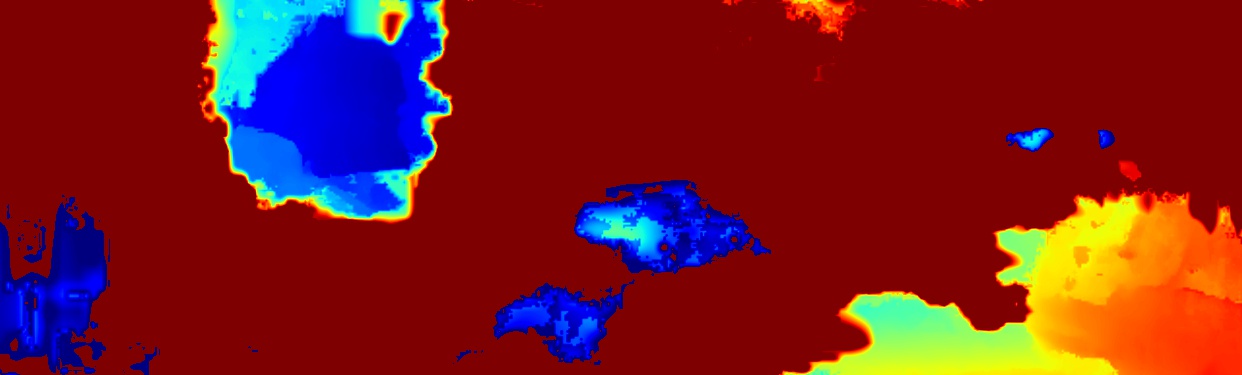}\\
			\includegraphics[width=1\linewidth]{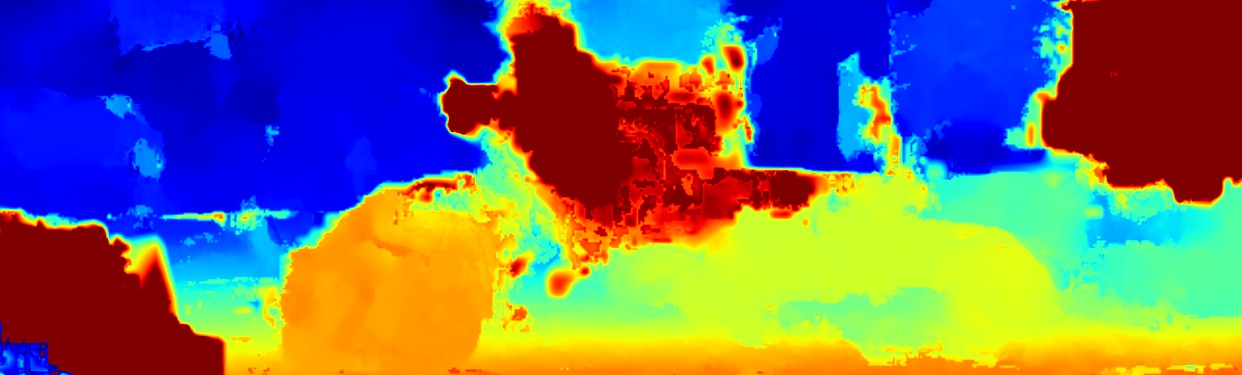}\\
			\includegraphics[width=1\linewidth,height=0.65\linewidth]{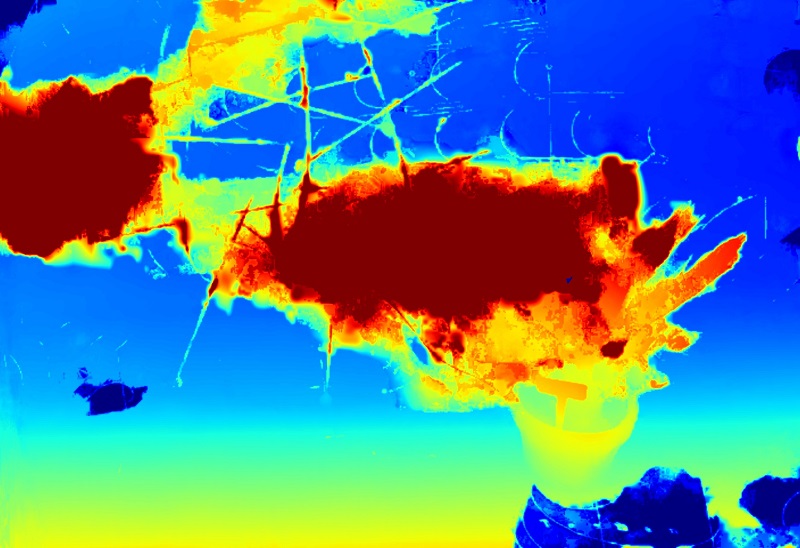}\\
			\includegraphics[width=1\linewidth,height=0.65\linewidth]{images/middle_hd3_2.jpg}\\
			\includegraphics[width=1\linewidth]{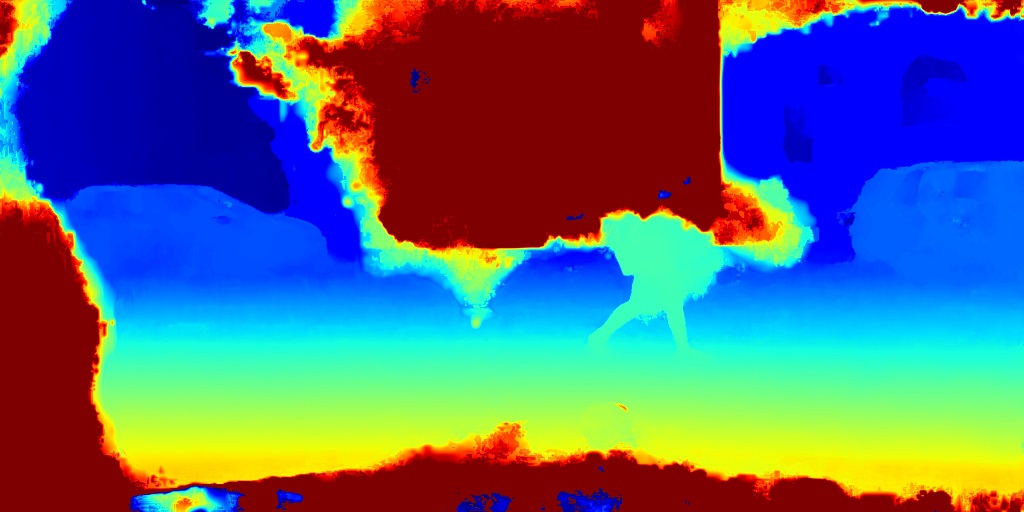}\\
			\includegraphics[width=1\linewidth]{images/city_hd3_22.jpg}
			\includegraphics[width=1\linewidth,height=0.55\linewidth]{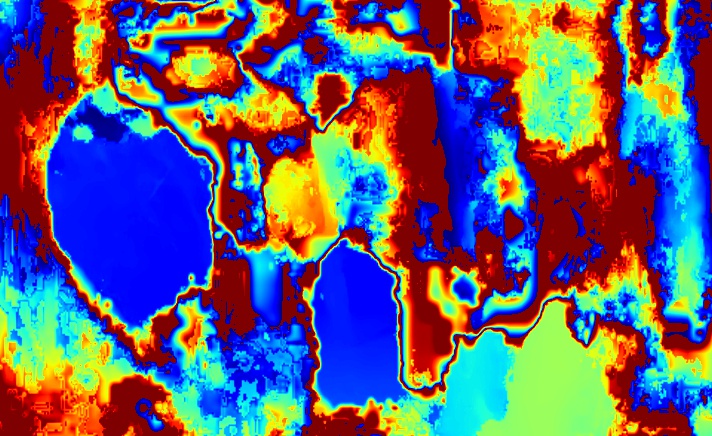}\\
			\includegraphics[width=1\linewidth,height=0.55\linewidth]{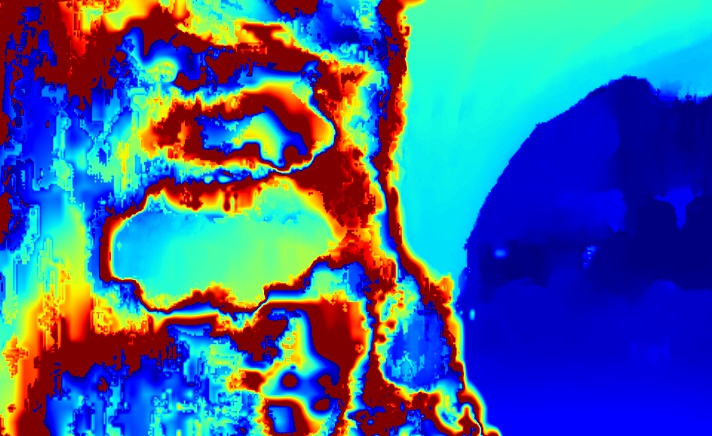}
		\end{minipage}
	}
	\hspace{-3mm}
	\subfigure[PSMNet\cite{chang2018pyramid}]{
		\begin{minipage}[b]{0.247\linewidth}
			\includegraphics[width=1\linewidth]{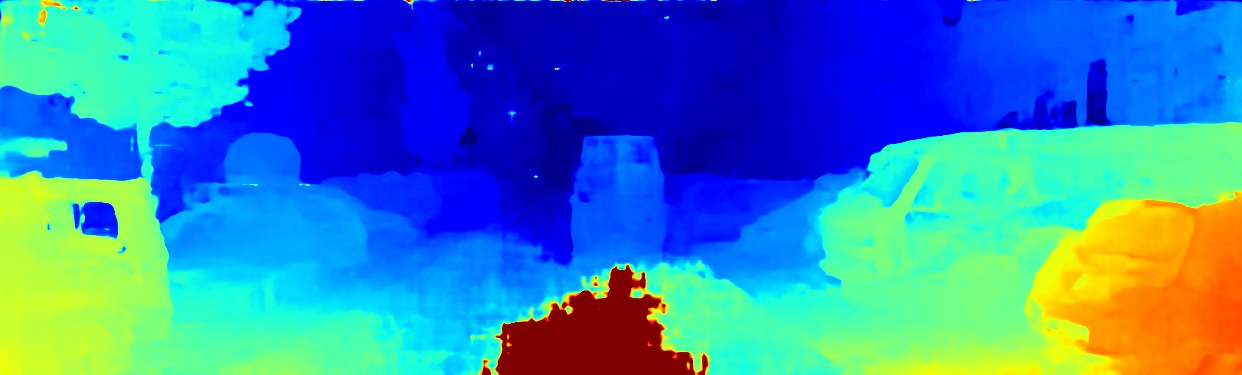}\\
			\includegraphics[width=1\linewidth]{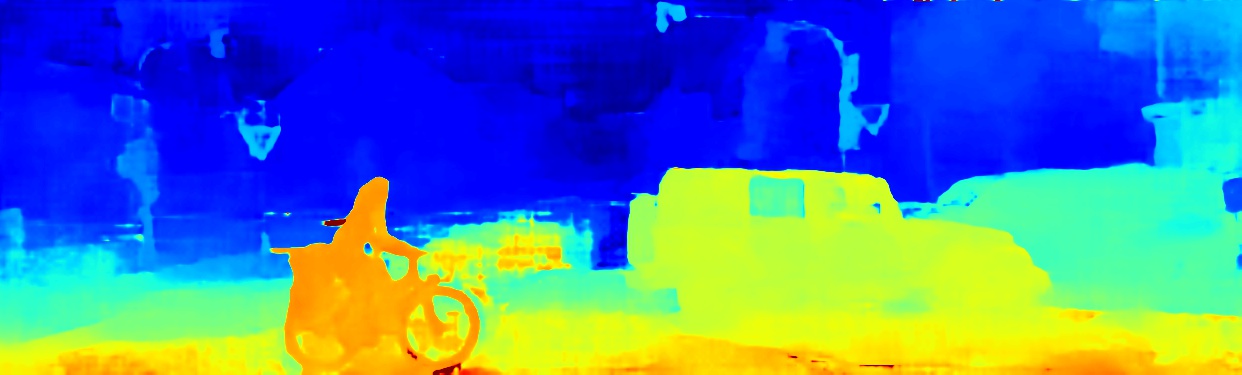}\\
			\includegraphics[width=1\linewidth,height=0.65\linewidth]{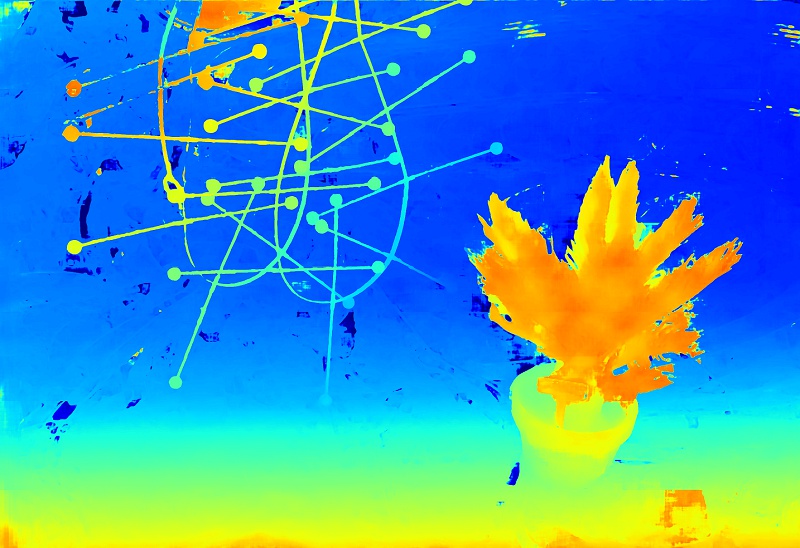}\\
			\includegraphics[width=1\linewidth,height=0.65\linewidth]{images/middle_psmnet_2.jpg}\\
			\includegraphics[width=1\linewidth]{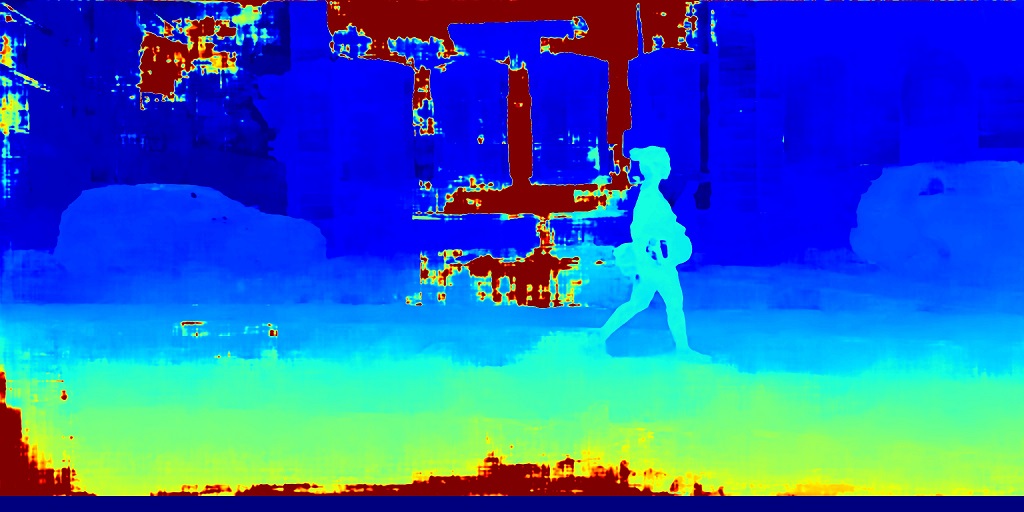}\\
			\includegraphics[width=1\linewidth]{images/city_psmnet_22.jpg}
			\includegraphics[width=1\linewidth,height=0.55\linewidth]{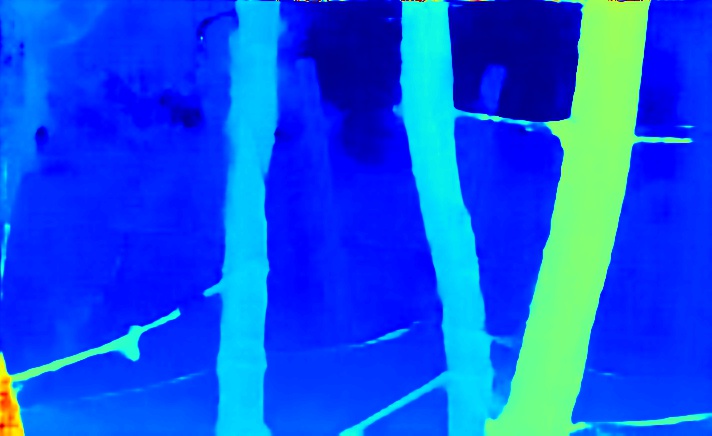}\\
			\includegraphics[width=1\linewidth,height=0.55\linewidth]{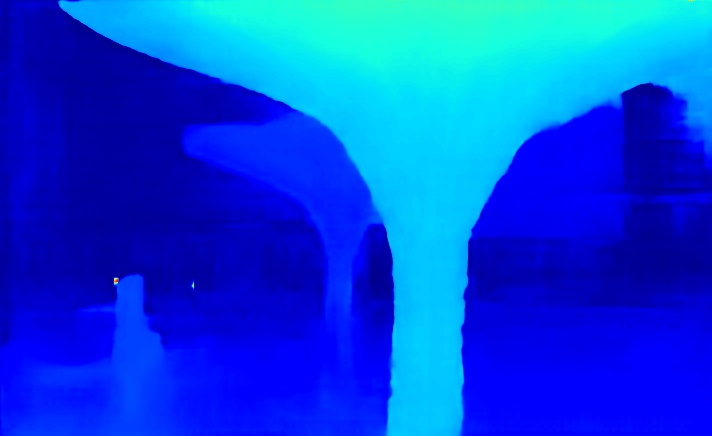}
		\end{minipage}
	}
	\hspace{-3mm}
	\subfigure[Our DSMNet]{
		\begin{minipage}[b]{0.247\linewidth}
			\includegraphics[width=1\linewidth]{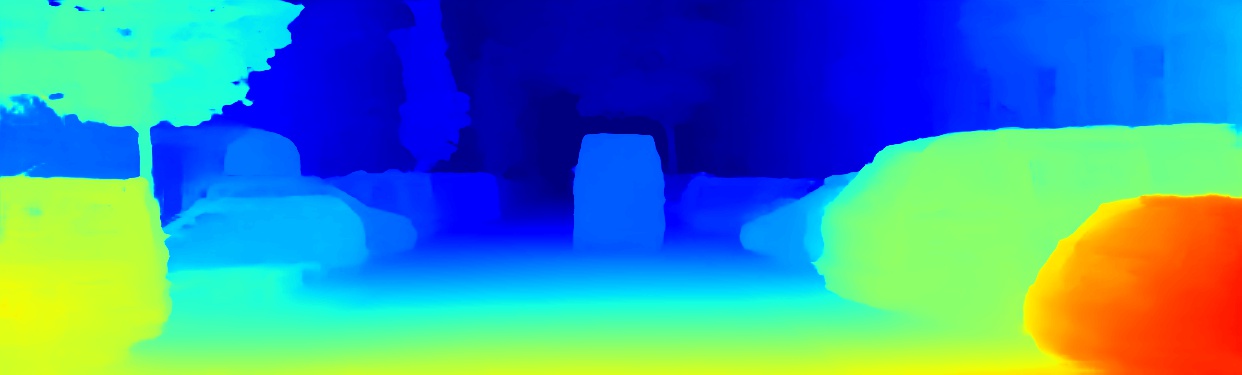}\\
			\includegraphics[width=1\linewidth]{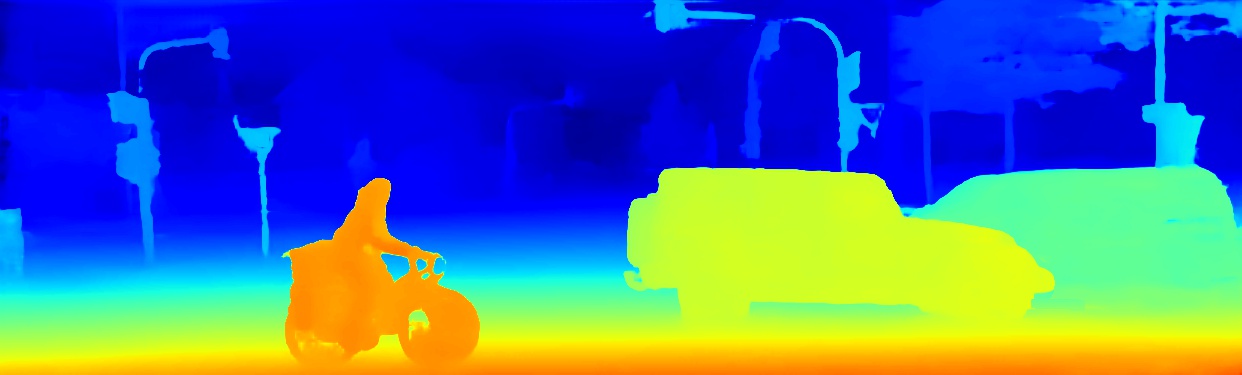}\\
			\includegraphics[width=1\linewidth,height=0.65\linewidth]{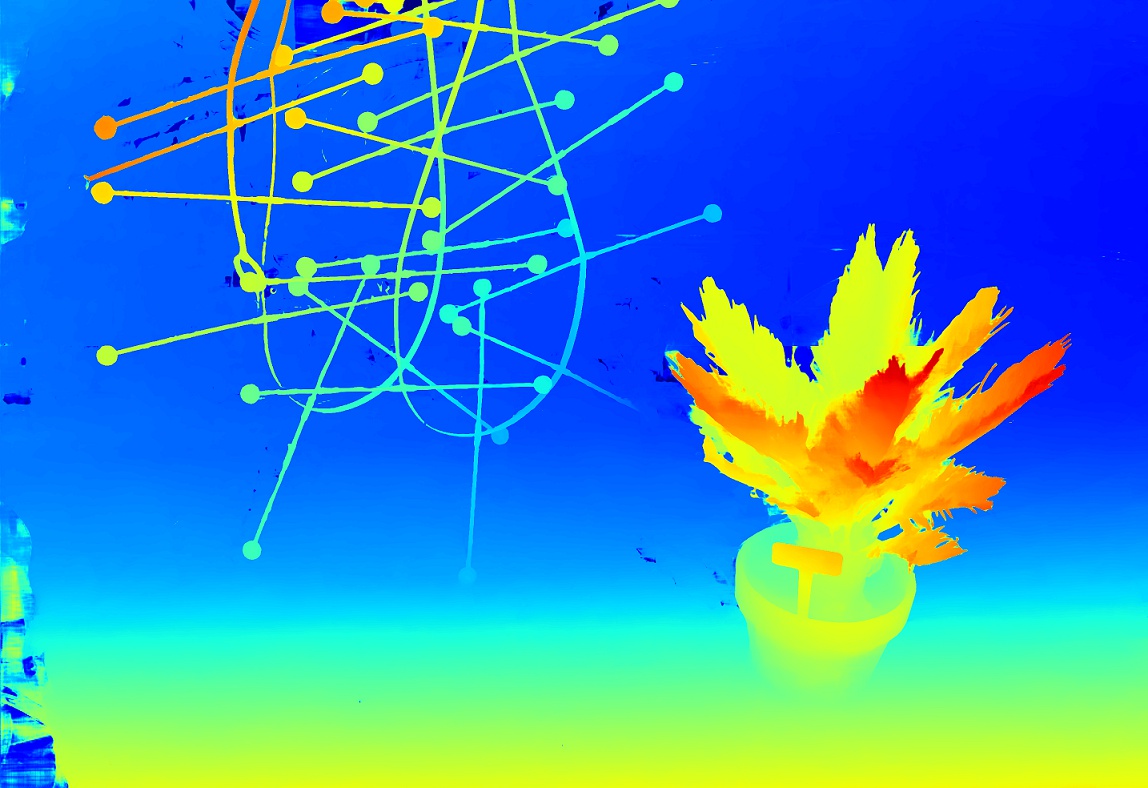}\\
			\includegraphics[width=1\linewidth,height=0.65\linewidth]{images/middle_dsmnet_2.jpg}\\
			\includegraphics[width=1\linewidth]{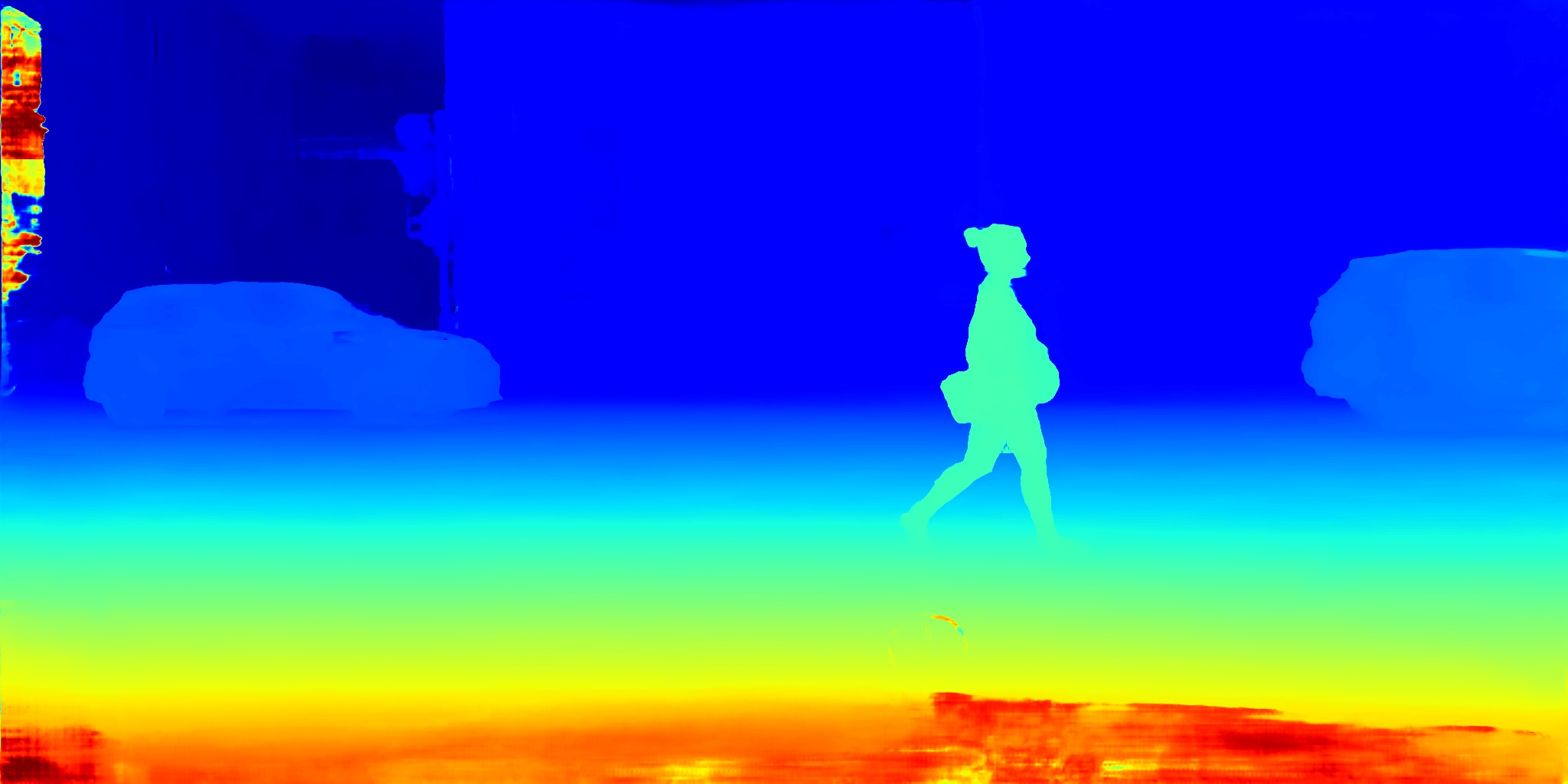}\\
			\includegraphics[width=1\linewidth]{images/city_dsmnet_22.jpg}
			\includegraphics[width=1\linewidth,height=0.55\linewidth]{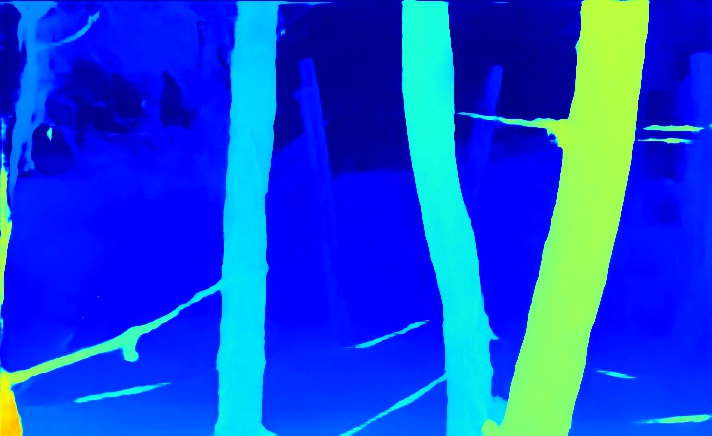}\\
			\includegraphics[width=1\linewidth,height=0.55\linewidth]{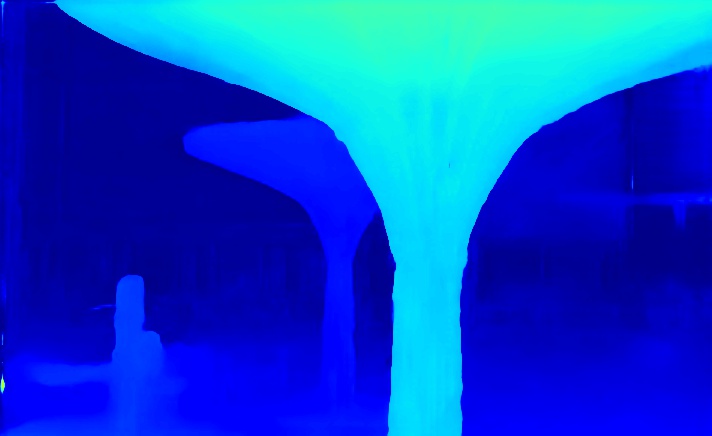}
		\end{minipage}
	}
	\caption{\small Comparisons with the state-of-the-art models on four real dataset (from top to bottom: KITTI, Middlebury, ETH3D and Cityscapes). All the models are trained on the synthetic dataset. Our DSMNet can produce accurate disparity estimation on other new datasets without fine-tuning.}
	\label{fig:resultsmore}
\end{figure*}

\subsection{Comparisons with Other Non-local Strategies}
Our graph-based filtering strategy is better for capturing the structural and geometric context for robust domain-invariant stereo matching. The non-local neural network denoising \cite{xie2019feature} and non-local attention \cite{huang2019ccnet} do not have spatial constraints that usually lead to over smoothness of the depth edges and thin structures (as shown in Fig.~\ref{fig:smooth}).
\begin{figure*}[t]
	\setlength{\abovecaptionskip}{-3pt}
	\setlength{\belowcaptionskip}{-10pt}
	\centering
	\vspace{-2mm}
	\begin{minipage}[b]{0.49\linewidth}
		\includegraphics[width=1\linewidth]{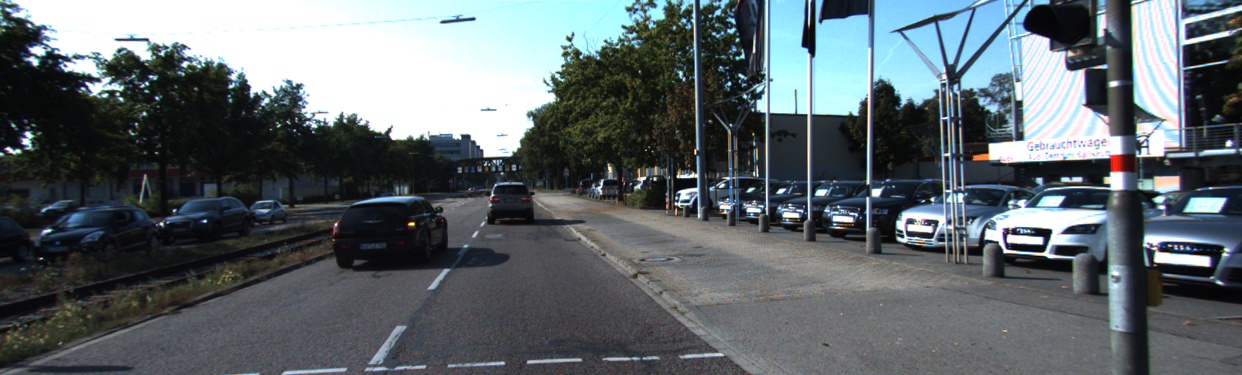}\\
		\includegraphics[width=1\linewidth]{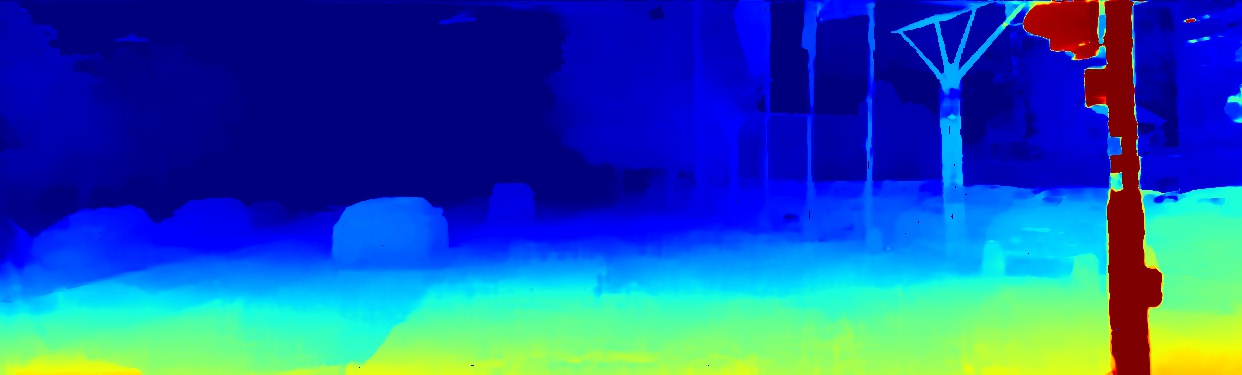}\\
		\includegraphics[width=1\linewidth]{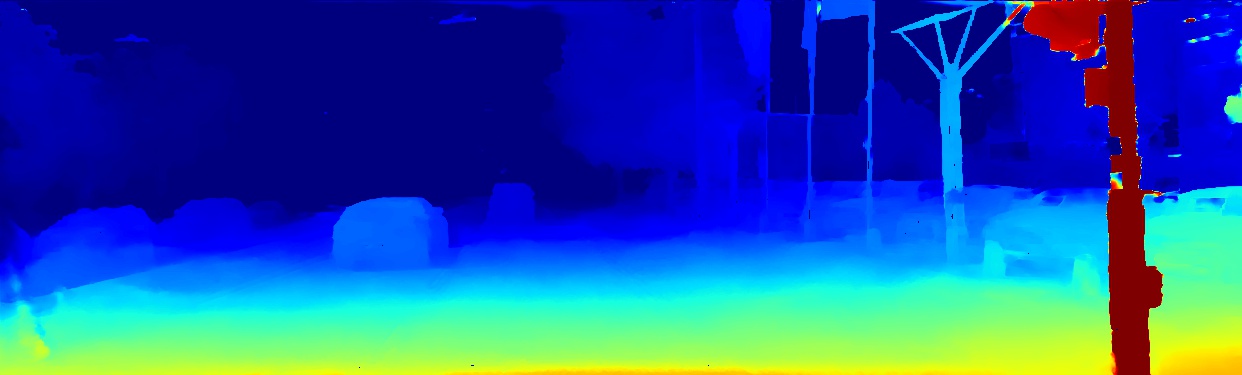}\\
		\includegraphics[width=1\linewidth]{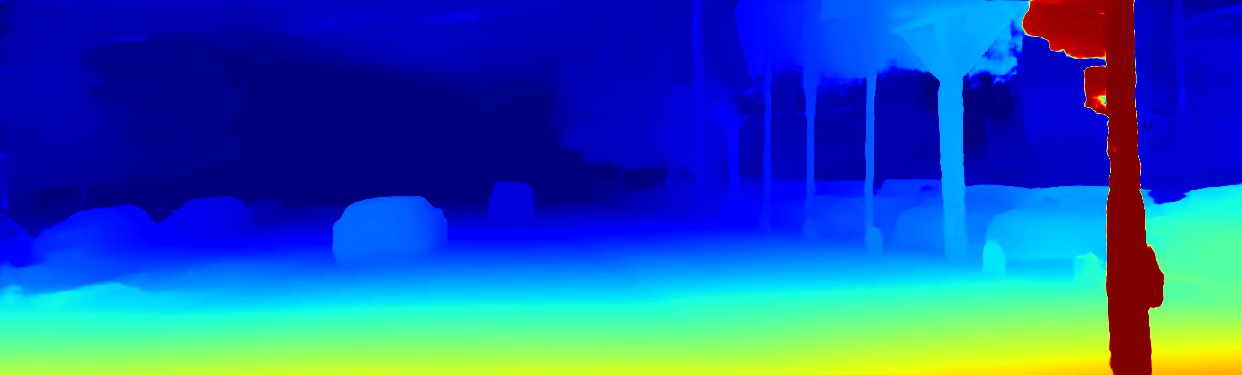}\\
	\end{minipage}
	\hspace{-0.5mm}
	\begin{minipage}[b]{0.49\linewidth}
		\includegraphics[width=1\linewidth]{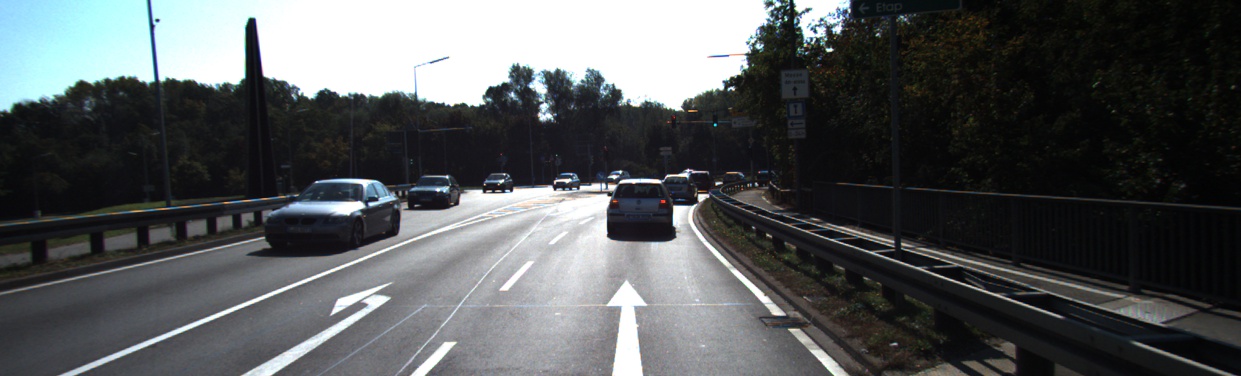}\\
		\includegraphics[width=1\linewidth]{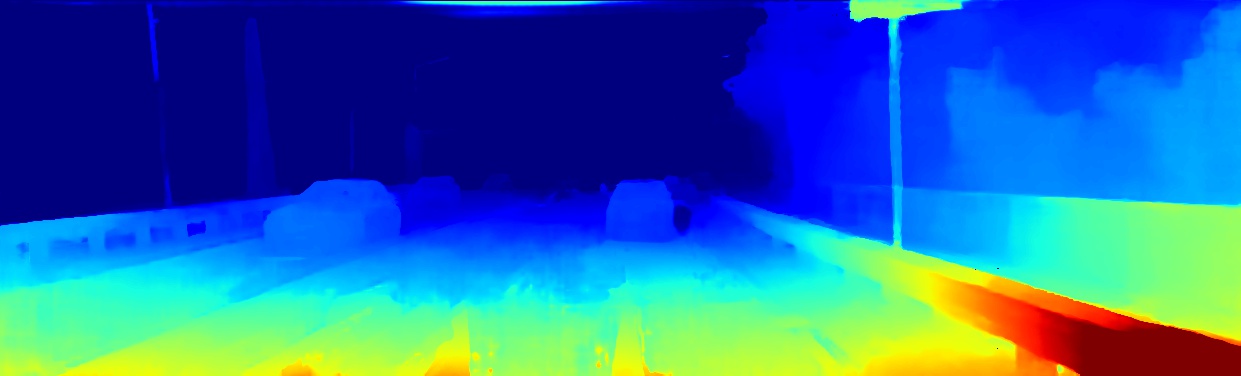}\\
		\includegraphics[width=1\linewidth]{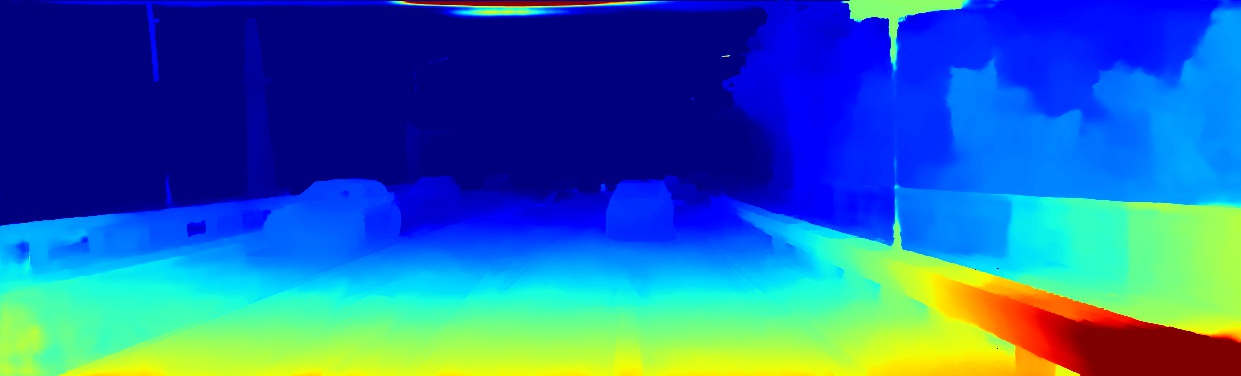}\\
		\includegraphics[width=1\linewidth]{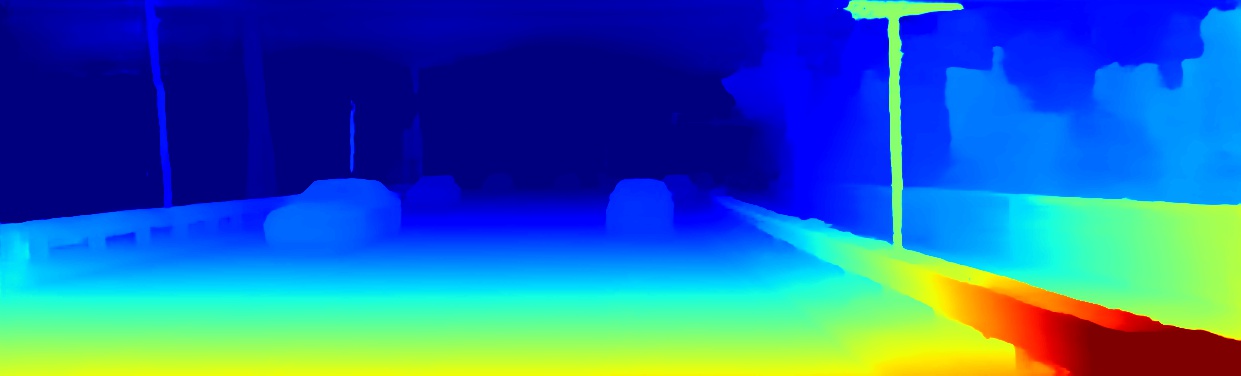}\\
	\end{minipage}
	\caption{\small Comparisons with non-local attention mechanism \cite{huang2019ccnet} ({\em second row}) and non-local denoising \cite{xie2019feature} strategy ({\em third row}). When using these strategies, the thin structures (\eg poles) are easily eroded by the background.  These non-local strategies easily smooth out the disparity maps. As a comparison, our DSMNet ({\em last row}) can keep the thin structures of the disparity maps.}
	\label{fig:smooth}
\end{figure*}

\end{document}